\documentclass[10pt,twocolumn,letterpaper]{article}

\usepackage[pagenumbers]{cvpr} 
\usepackage{times}
\usepackage{array}
\usepackage{epsfig}
\usepackage{amsmath}
\usepackage{amssymb}
\usepackage{booktabs}
\usepackage{multirow}
\usepackage{tabularx}
\usepackage{graphicx}
\usepackage{makecell}
\usepackage{cuted}
\usepackage{capt-of}
\usepackage[table]{xcolor}
\usepackage{comment}

\definecolor{cvprblue}{rgb}{0.21,0.49,0.74}
\newcommand{\bu}[1]{{\color{cvprblue}{#1}}}
\usepackage[pagebackref,breaklinks,colorlinks,allcolors=cvprblue]{hyperref}

\newcolumntype{L}[1]{>{\raggedright\arraybackslash}m{#1}}
\newcolumntype{C}[1]{>{\centering\arraybackslash}m{#1}}
\newcolumntype{R}[1]{>{\raggedleft\arraybackslash}m{#1}}
\newcolumntype{+}{>{\global\let\currentrowstyle\relax}}
\newcolumntype{^}{>{\currentrowstyle}}

\usepackage[capitalize]{cleveref}
\crefname{section}{Sec.}{Secs.}
\Crefname{section}{Section}{Sections}
\crefname{figure}{Fig.}{Figs.}
\Crefname{figure}{Figure}{Figures}
\crefname{table}{Tab.}{Tabs.}
\Crefname{table}{Table}{Tables}

\def\paperID{31756} 
\def\confName{CVPR}
\def\confYear{2026}

\title{CanonCGT: Reference-Based Color Grading via Canonical Pivot Representation}

\author{Jinwon Ko\\
Korea University\\
{\tt\small jwko@mcl.korea.ac.kr}
\and
Keunsoo Ko\\
The Catholic University of Korea\\
{\tt\small ksko@catholic.ac.kr}
\and
Chang-Su Kim\thanks{Corresponding author.}\\
Korea University\\
{\tt\small changsukim@korea.ac.kr}}

\begin{document}
\maketitle

\begin{strip}
\centering
\vspace*{-1.75cm}
\includegraphics[width=\textwidth]{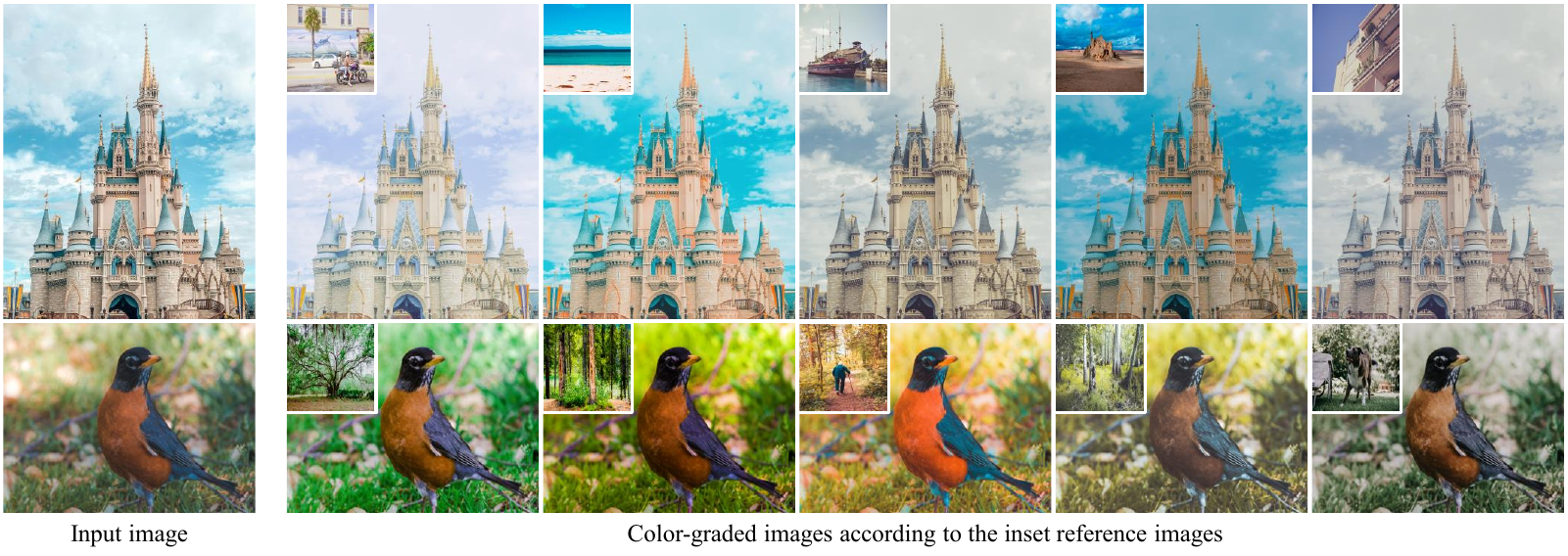}
\vspace{-0.6cm}
\captionof{figure}{\textbf{Our reference-based color grading results.} Each input image is color-graded to match the tonal mood and lighting of its reference, yielding photorealistic results that preserve color harmony and scene structure.}
\label{fig:Intro}
\end{strip}

\begin{abstract}
Reference-based color grading aims to reproduce the tonal mood and lighting of a reference while preserving color harmony and scene structure. Existing photorealistic and filter-based methods often produce unstable tone mappings --- over-shifting or inconsistently retaining colors --- leading to unnatural results. We propose CanonCGT, a two-stage framework built on a canonical pivot --- a style-neutral intermediate representation for stable color mapping. The first stage canonicalizes the input by removing intrinsic tonal bias, and the second color-grades it to match the reference style. A dual-phase training scheme, DP-CGT, combines supervised preset learning with self-supervised refinement on unpaired photographs. CanonCGT delivers photorealistic and tonally consistent results across diverse datasets, surpassing state-of-the-art methods in stability and visual fidelity. Our codes are available at \href{https://github.com/Jinwon-Ko/CanonCGT}{https://github.com/Jinwon-Ko/CanonCGT}.
\end{abstract}

\begin{figure*}[!t]
\centering
\includegraphics[width=1\linewidth]{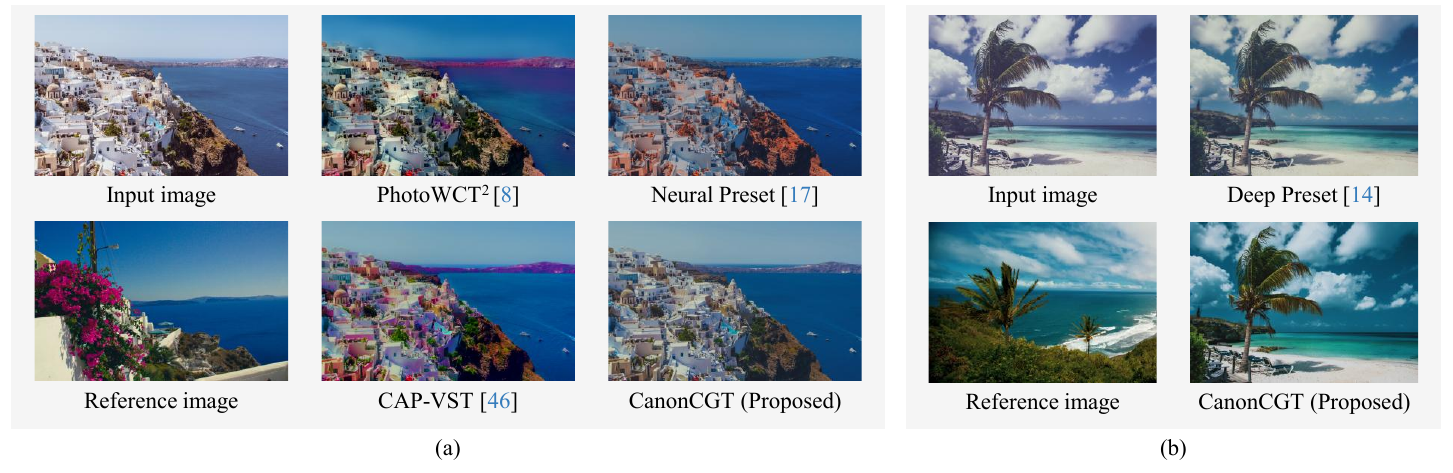}
\vspace*{-0.7cm}
\caption{Color grading results of existing methods and the proposed CanonCGT. (a) Comparison with recent photorealistic style transfer methods \cite{chiu2022photowct2,ke2023neural,wen2023cap}, which often cause excessive tonal shifts and texture distortions. (b) Comparison with a filter-based approach \cite{ho2021deep}, which tends to retain or accumulate existing stylistic effects instead of reproducing the reference tone. In contrast, CanonCGT faithfully transfers the reference color grading while preserving structural fidelity and tonal stability.}
\vspace*{-0.3cm}
\label{fig:Related}
\end{figure*}

\vspace{-0.4cm}
\section{Introduction}
\vspace{-0.1cm}

Color grading --- the process of adjusting color, tone, and overall mood to achieve a desired visual impression --- is central to professional photo editing and cinematography \cite{magrin2022colorgrading,zhang2025colorgrading}. It enhances atmosphere, maintains stylistic coherence, and ensures visual balance through subtle tonal adjustments. Manual color grading, however, requires expertise and time, as users must precisely tune parameters such as exposure, contrast, and color temperature. Modern tools offer these controls but remain difficult for non-experts.

To simplify this process, many applications provide predefined filters or lookup tables (LUTs) that emulate specific photographic moods. While convenient, these presets often fail to reflect the user's intended aesthetic or scene characteristics and yield inconsistent outcomes because they ignore contextual or pre-existing tonal bias.

A more intuitive solution is reference-based color grading, which allows users to specify a desired aesthetic by providing a reference photo. As illustrated in \Cref{fig:Intro}, this approach reproduces the reference’s tonal mood and lighting without manual tuning, offering an accessible framework for personalized photo enhancement.

Recent photorealistic style transfer methods have advanced this goal, following neural style transfer \cite{gatys2016image}, which first used deep features to represent visual styles. Later approaches \cite{li2018closed,yoo2019photorealistic,chiu2022photowct2,ke2023neural} transfer styles in feature space by aligning statistics or modulating activations. However, they often produce exaggerated tonal shifts and texture distortions, as shown in \Cref{fig:Related}(a), making them unsuitable for color grading, which demands subtle tonal control and structural fidelity.

Filter style transfer \cite{yim2020filter,ho2021deep} learns subtle, retouching-oriented tone adjustments from paired datasets of natural and filtered images. By modeling low-level color mappings, these methods produce coherent, visually faithful results. However, their mappings remain limited to natural-to-filtered pairs --- when the input already shows a strong stylistic bias, they often retain the original tone instead of adaptively restyling it toward the reference, as shown in \Cref{fig:Related}(b). This limitation calls for a mechanism that neutralizes pre-existing styles and establishes a canonical representation as a stable basis for reference-based color grading.

In this paper, we propose CanonCGT, a reference-based color grading algorithm that addresses these limitations through a canonical pivot --- a style-neutral intermediate representation providing a stable basis for adaptive color mapping. CanonCGT operates in two stages: (1) canonicalization, which maps the input to the canonical domain to remove intrinsic tonal bias, and (2) grading, which transfers the reference tone onto this representation.

To train the two-stage pipeline, we develop dual-phase color grading training (DP-CGT). The first, supervised phase learns reliable tone mappings across diverse presets by training on multiple filtered versions of each canonical image. The second, self-supervised phase refines CanonCGT on unconstrained photographs by reconstructing locally perturbed regions using nearby grading cues, enabling robust and general color grading beyond preset-defined styles. Extensive experiments show that CanonCGT achieves superior photorealism and generalization over state-of-the-art photorealistic and filter-based methods.

This paper makes the following main contributions:

\begin{itemize}
    \item We introduce the concept of a \emph{canonical pivot} --- a style-neutral representation that provides a stable intermediate domain for reference-based color grading.
    \item We design a two-stage \emph{canonicalization-grading} framework that removes tonal bias and applies reference styles in a decoupled yet complementary manner.
    \item We develop \emph{DP-CGT}, a dual-phase training strategy combining supervised preset learning with self-supervised generalization for robust and adaptive color grading.
    \item \emph{CanonCGT} surpasses recent photorealistic and filter-based methods \cite{an2020ultrafast,ho2021deep,chiu2022photowct2,ke2023neural,wen2023cap}, consistently achieving high photorealism and tonal stability across datasets.
\end{itemize}

\begin{figure*}[!t]
\centering
\includegraphics[width=1\linewidth]{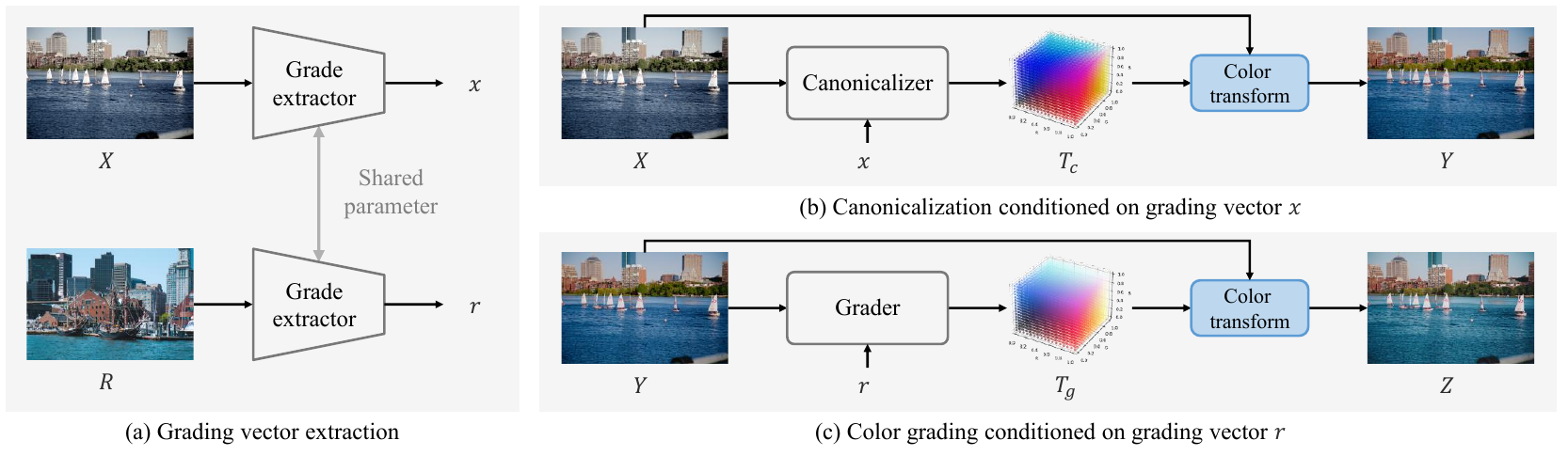}
\vspace*{-0.7cm}
\caption{Overview of the CanonCGT framework. 
    The grade extractor derives grading vectors $x$ and $r$ from the input $X$ and reference $R$. 
    The canonicalizer and grader generate 3D LUTs $T_c$ and $T_g$ to produce the canonical image $Y$ and retouched output $Z$, respectively.}
\label{fig:Overview}
\vspace*{-0.3cm}
\end{figure*}

\section{Related Work}

\subsection{Style transfer}
\noindent\textbf{Artistic style transfer:} Early color transfer methods matched global color statistics between images \cite{reinhard2002color,pitie2005n,pitie2007automated}, but deep learning later enabled more expressive style manipulation. Neural style transfer \cite{gatys2016image} reproduced artistic effects on natural photos and inspired numerous follow-up studies \cite{chen2017stylebank,liu2021adaattn,zhang2022domain,dumoulin2016learned,deng2022stytr2}, emphasizing painterly abstraction --- effective for artistic creation but undesirable for photographic editing --- often yielding unrealistic color reproduction and structural distortion.

\medskip
\noindent\textbf{Photorealistic style transfer:} Photorealistic style transfer modifies tone and color while preserving scene structure by aligning feature statistics between content and reference images in latent space. Li \etal \cite{li2017universal,li2018closed} and Yoo \etal \cite{yoo2019photorealistic} employed the whitening-and-coloring transform (WCT) to match feature distributions, while An \etal \cite{an2020ultrafast} and Chiu \etal \cite{chiu2022photowct2} improved efficiency through neural architecture search and block-wise training. Wen \etal \cite{wen2023cap} enhanced spatial coherence via content-affinity preservation, and Ke \etal \cite{ke2023neural} introduced a neural-preset model that disentangles content and style.

Despite these advances, photorealistic style transfer often causes excessive tonal shifts or local color bleeding, as shown in Figure~\ref{fig:Related}(a), hindering the precise tone and illumination control required in photographic color grading.

\medskip
\noindent\textbf{Filter style transfer and color grading:} Recent filter style transfer methods \cite{yim2020filter,ho2021deep} learn low-level color mappings from paired datasets of natural and filtered images to emulate predefined retouching styles through supervised learning. Yim \etal \cite{yim2020filter} trained parametric transformations for tone adaptation, and Ho \etal \cite{ho2021deep} unified multiple styles within a single network. Although these methods produce visually pleasing results, their mappings remain limited to the filter sets used for training and fail to generalize to unseen references. When applied to already edited images, they often adapt poorly --- either retaining the original tone or accumulating existing effects --- leading to unbalanced results, as shown in \Cref{fig:Related}(b).

To address these limitations, we propose CanonCGT, which leverages canonical pivot images to stabilize reference-based tone mapping and suppress residual or accumulated styles. CanonCGT reproduces the desired photographic mood and lighting of a reference --- without structural deformation, tonal exaggeration, or style accumulation --- effectively overcoming the main drawbacks of artistic, photorealistic, and filter style transfer methods.

\subsection{LUT-based color mapping}
A LUT represents a color mapping compactly and interpretably by defining explicit correspondences between input and output colors. Unlike implicit feature-space transformations, LUTs provide deterministic and spatially consistent tone adjustments, making them well suited for color grading, where global balance and structural realism are crucial \cite{karaimer2016software}.

Recent studies have exploited these advantages for photorealistic enhancement. Zeng \etal \cite{zeng2020learning} predicted image-adaptive LUTs for real-time enhancement. Yang \etal \cite{yang2022adaint,yang2022seplut} improved flexibility and efficiency through adaptive and separable LUTs. Ko \etal \cite{ko2025lutformer} employed transformer architectures to predict structured 3D LUTs for image-adaptive color mapping. Other methods integrated LUTs into reference-guided style transfer \cite{lin2023adacm,chen2023nlut,li2025dlut}, achieving stable, structure-preserving color adaptation. More recently, Shin \etal \cite{shin2025video} extended LUT-based color grading to video by generating a diffusion-driven LUT from key-frames. Building on these advances, the proposed CanonCGT employs LUT-based transformations for both canonicalization and grading, ensuring reliable tone mapping with high structural fidelity.

\begin{figure*}[!t]
\centering
\includegraphics[width=1\linewidth]{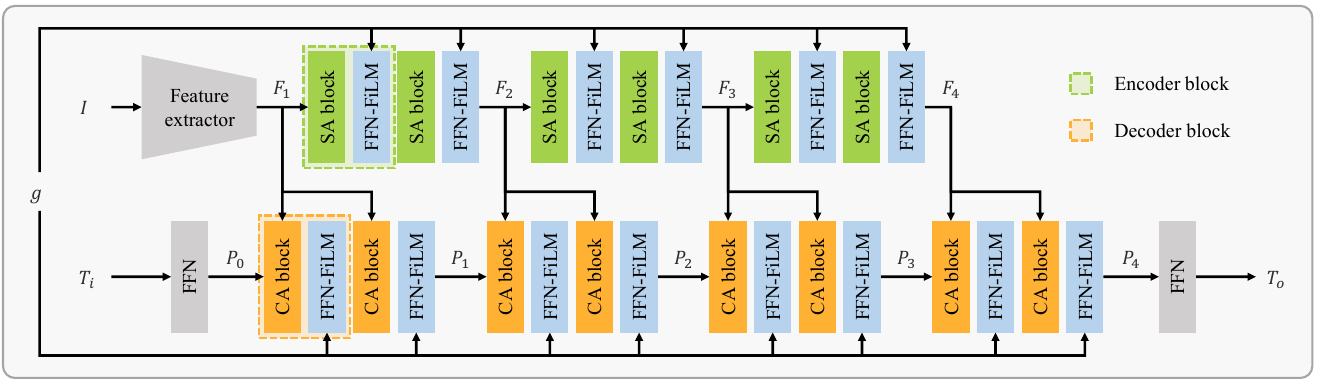}
\caption{The architecture of the canonicalizer and grader. The network, referred to as the conditioned LUT generator, takes an input image $I$ and a condition vector $g$ to modulate the identity LUT $T_i$, producing an image-adaptive LUT $T_o$ for color transformation. The vector $g$ denotes the input grade $x$ when used for canonicalization and the reference grade $r$ when used for stylization.}
\label{fig:Backbone}
\vspace*{-0.2cm}
\end{figure*}

\section{Proposed Algorithm}
\label{sec:proposed}

Most photographs undergo in-camera or post-capture processing, and many are further edited with presets or filters. We collectively refer to these transformations as a grading style, which defines the overall tone and mood of an image. To achieve stable reference-based color grading without style retention or accumulation, we employ a canonicalization-grading framework, as shown in \Cref{fig:Overview}.

CanonCGT adjusts the color grading of an input image $X$ to match that of a reference $R$. A grade extractor derives grading vectors $x$ and $r$ from $X$ and $R$, respectively, where a grade denotes a grading vector for brevity. The canonicalizer removes the latent grade $x$ from $X$ to produce a style-neutral canonical image $Y$, and the grader applies the reference grade $r$ to $Y$, generating the retouched output $Z$. Both modules employ LUT-based color transforms for spatially consistent and structure-preserving tone adaptation.

\subsection{CanonCGT}

\noindent\textbf{Grade extractor:}
The grade extractor encodes the tonal characteristics of an image into a latent representation that defines its grading style. The extracted grade conditions the canonicalizer to remove intrinsic style bias or the grader to impose the reference style. A pretrained MobileNet-v2 \cite{sandler2018mobilenetv2} serves as a lightweight backbone.

Since grading styles are diverse and unlabeled, the embedding space is organized through contrastive learning \cite{khosla2020supervised} using known presets: images edited with the same preset form positive pairs, while those with different presets form negative ones. Each preset defines a consistent tonal transformation, enabling the grade extractor to learn a compact, tone-aware latent space, as shown in the supplement.

\medskip
\noindent\textbf{Canonicalizer:}
The canonicalizer predicts a LUT to remove intrinsic style bias by modeling local color and global tonal context. As shown in \Cref{fig:Backbone}, it takes the input image $I$ and its grading vector $g$ to extract multi-level features that capture spatial and chromatic dependencies. These features are modulated by $g$ through FiLM \cite{perez2018film} layers and aggregated to produce the LUT.

Given a resized input $I \in \mathbb{R}^{L \times L \times 3}$, a CNN-based feature extractor produces a feature map $F \in \mathbb{R}^{\frac{L}{8} \times \frac{L}{8} \times C}$ encoding local color and texture. The map is flattened into a token sequence $F_1 \in \mathbb{R}^{\frac{L^2}{64} \times C}$ and processed by six encoder blocks that capture global dependencies and build multi-level representations $\{F_l\}_{l=1}^4$. 

Each encoder block consists of a self-attention (SA) block and an FFN-FiLM block, as illustrated in \Cref{fig:Backbone}. In the SA block, queries, keys, and values are linearly projected from $F_l$ and updated by multi-head self-attention \cite{vaswani2017attention} with a residual connection:
\begin{equation}
F_{l}^{(\mathrm{sa})} = \operatorname{MHSA}(F_l) + F_l,
\label{eq:sa_block}
\end{equation}
where $\operatorname{MHSA}(\cdot)$ captures long-range color dependencies within the feature representation.

\begin{figure}[!t]
\centering
\includegraphics[width=1\linewidth]{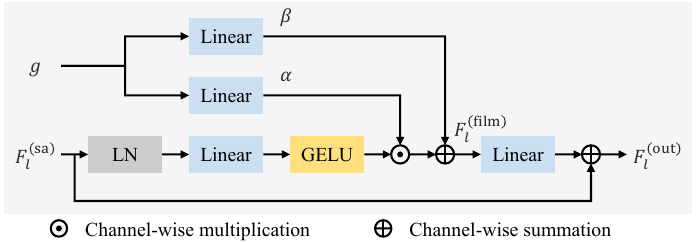}
\caption{The architecture of the FFN-FiLM block.}
\label{fig:FFN_FiLM}
\vspace*{-0.2cm}
\end{figure}

The features are then refined by the FFN-FiLM block, illustrated in \Cref{fig:FFN_FiLM}. Given the output $F_{l}^{(\mathrm{sa})}$ in \eqref{eq:sa_block} and the grading vector $g$, the block applies layer normalization and a feed-forward transform, followed by FiLM-based channel modulation:
\begin{align}
    \alpha &= W_1 g, \quad \beta = W_2 g, 
    \label{eq:film1} \\
    F_{l}^{(\mathrm{film})} &= \alpha \odot \sigma(F_{l}^{(\mathrm{sa})} W_3) + \beta, \label{eq:film2} \\
    F_{l}^{(\mathrm{out})} &= F_{l}^{(\mathrm{film})} W_4 + F_{l}^{(\mathrm{sa})}, \label{eq:film3}
\end{align}
where $\sigma(\cdot)$ denotes GELU \cite{hendrycks2016gelu}. The grading vector $g \in \mathbb{R}^{C}$ generates scale and shift parameters $\alpha, \beta \in \mathbb{R}^{C'}$, while $W_1, W_2 \in \mathbb{R}^{C'\times C}$, $W_3 \in \mathbb{R}^{C\times C'}$, and $W_4 \in \mathbb{R}^{C'\times C}$ are learnable projections. Unlike the original FiLM \cite{perez2018film} that conditions on language features, our FFN-FiLM uses a grading vector to modulate feature responses in color and tone. This conditioning is integrated into the FFN through projections ($W_3$, $W_4$) and a residual path, adapting FiLM for canonicalization or grading.

\begin{figure*}[!t]
\centering
\includegraphics[width=1\linewidth]{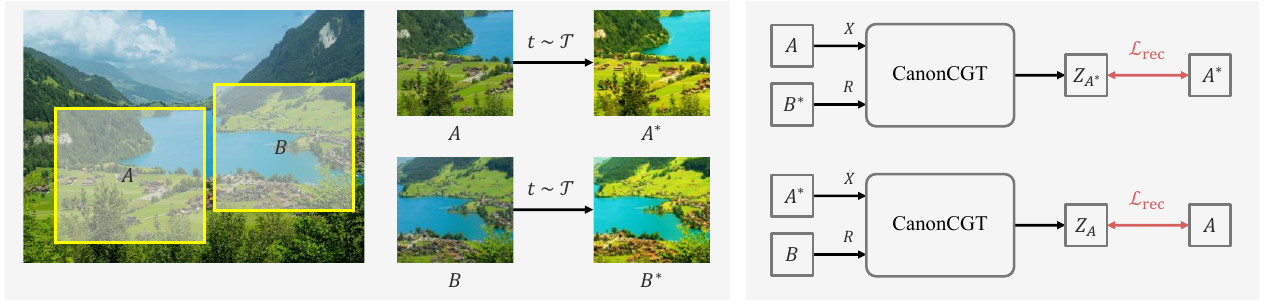}  
\vspace*{-0.5cm}
\caption{Self-supervised learning phase: Two crops $A$ and $B$ are randomly sampled from the same photo and perturbed by color grading transformations $t \sim \mathcal{T}$, producing $A^{\ast}$ and $B^{\ast}$. CanonCGT processes $(A, B^{\ast})$ and $(A^{\ast}, B)$ symmetrically, each optimized with a reconstruction loss $\mathcal{L}_{\mathrm{rec}}$.}
\label{fig:SSL}
\vspace*{-0.2cm}
\end{figure*}    

A second encoder block, conditioned on the same $g$, is then applied to $F_{l}^{(\mathrm{out})}$ to yield the next-level feature $F_{l+1}$. After extracting multi-level features $\{F_{l}\}_{l=1}^{4}$, we initialize the query tokens $P_0 \in \mathbb{R}^{N^{3}\times C}$ by projecting each grid point of the identity LUT $T_i \in \mathbb{R}^{N\times N\times N\times 3}$ through a lightweight FFN. These tokens are progressively refined through eight decoder blocks, each composed of a cross-attention (CA) block and an FFN-FiLM block. 

In the CA block, queries are projected from the LUT tokens $P_{l-1}$, while keys and values are projected from the encoded feature map $F_l$. The LUT tokens are updated by multi-head cross-attention \cite{vaswani2017attention}:
\begin{equation}
P_{l-1}^{(\mathrm{ca})} = \operatorname{MHCA}(P_{l-1}, F_l) + P_{l-1},
\end{equation}
where $\operatorname{MHCA}(\cdot)$ captures scene-dependent color correlations between the LUT tokens and image features. This operation enables each LUT entry to incorporate contextual cues from the encoded representation, adapting its color mapping to the scene’s tonal characteristics.

The FFN-FiLM block modulates $P_{l-1}^{(\mathrm{ca})}$ with the grade $g$ to produce an intermediate $P_{l-1}^{(\mathrm{out})}$, as in  \eqref{eq:film1}$\sim$\eqref{eq:film3}. A second decoder block, conditioned on the same $F_l$ and $g$, is applied to $P_{l-1}^{(\mathrm{out})}$ to obtain the next-level output $P_l$. After four such levels (eight decoder blocks in total), the refined tokens $P_4$ are projected through an additional FFN to generate the output LUT $T_o$. When operating as a canonicalizer, $T_o$ becomes the canonical LUT $T_c$, which is applied to $X$ to produce the canonical image $Y$.

\medskip
\noindent\textbf{Grader:}
The grader employs the same architecture as the canonicalizer but is conditioned on the reference grade $r$. Conceptually, the canonicalizer performs a many-to-one mapping that normalizes diverse input styles into a canonical representation, while the grader performs a one-to-many mapping that transforms the canonical style into target styles. Given the canonical image $Y$, the grader predicts a grading LUT $T_g$, which is applied to $Y$ to produce the final output $Z$, imparting the tonal and chromatic characteristics of the reference.

\begin{table*}[t]
\vspace*{-0.1cm}
\centering
\footnotesize
\caption{Quantitative comparison of color grading results (average performance over all test sets). *Results reproduced by our implementation, as official code is unavailable. The best and second-best results are shown in \textbf{bold} and \underline{underline}, respectively.}
\label{table:Quantitative}
\vspace*{-0.2cm}
\begin{tabular}[t]{lcccccccc}
\toprule
& \multicolumn{3}{c}{Fidelity} & \multicolumn{2}{c}{Content preservation} & \multicolumn{2}{c}{Grading style alignment} \\
\cmidrule(l){2-4} \cmidrule(l){5-6} \cmidrule(l){7-8}
Method & PSNR $\uparrow$ & SSIM $\uparrow$ & $\Delta E_{ab} \downarrow$ & LPIPS $\downarrow$ & SSIM$_\textrm{ED}$ $\uparrow$ & H-Corr $\uparrow$ & H-Chi $\downarrow$ \\
\midrule
PhotoNAS~\cite{an2020ultrafast}        & 16.71 & 0.7436 & 19.40 & 0.3174 & 0.6420 & 0.2547 & 0.3301 \\
PhotoWCT$^{2}$~\cite{chiu2022photowct2} & 16.36 & 0.8130 & 19.96 & 0.2256 & 0.7342 & 0.3207 & \underline{0.2869} \\
Neural Preset$^{\ast}$~\cite{ke2023neural} & 18.50 & 0.8451 & 17.28 & 0.2226 & 0.7174 & 0.2783 & 0.3526 \\
CAP-VST~\cite{wen2023cap}              & 18.00 & 0.8058 & 18.60 & 0.2335 & 0.7112 & \underline{0.3249} & 0.2924 \\
Deep Preset~\cite{ho2021deep}          & \underline{18.62} & \underline{0.8582} & \underline{15.21} & \underline{0.1750} & \underline{0.7575} & 0.2752 & 0.3185 \\
\midrule
\textbf{CanonCGT}                      & \textbf{28.99} & \textbf{0.9608} & \textbf{5.46} & \textbf{0.0665} & \textbf{0.8933} & \textbf{0.5204} & \textbf{0.1785} \\
\bottomrule
\end{tabular}
\vspace*{-0.1cm}
\end{table*}

\subsection{DP-CGT}
\label{subsec:DP-CGT}

CanonCGT is trained with DP-CGT, a dual-phase scheme consisting of a supervised warm-up phase and a self-supervised generalization phase. The first phase uses the MIT-Adobe FiveK dataset \cite{bychkovsky2011learning}, where expert~C retouchings define the canonical style. Fifty-six additional styles synthesized with Lightroom presets introduce diverse tonal and chromatic variations, enabling CanonCGT to learn reliable LUT-based transformations. The second phase refines the model on large-scale photographic collections for broader generalization to real-world grading styles.

\medskip
\noindent\textbf{Supervised phase:}
CanonCGT learns to translate images across preset styles while preserving tonal fidelity through three steps. First, the grade extractor is trained with the supervised contrastive loss $\mathcal{L}_{\mathrm{supcon}}$ \cite{khosla2020supervised} to form a discriminative embedding space. Second, with the extractor frozen, the canonicalizer and grader are trained using reconstruction losses $\mathcal{L}_{\mathrm{rec}}$ that enforce pixel, gradient, and perceptual fidelity --- between the input and canonical image for the canonicalizer, and between the graded output and ground truth for the grader. Finally, all modules are jointly fine-tuned end-to-end under the combined objectives, consolidating stable canonical mapping and tone reconstruction.

\medskip
\noindent\textbf{Self-supervised phase:}
CanonCGT is further refined on unlabeled photographs to improve generalization. Without style supervision, the grade extractor is frozen, and the canonicalizer and grader are optimized through self-referential reconstruction. As shown in \Cref{fig:SSL}, two crops $A$ and $B$ are randomly sampled from the same image and perturbed by color grading transformations $t \sim \mathcal{T}$, producing $A^{\ast}$ and $B^{\ast}$. CanonCGT processes $(A, B^{\ast})$ and $(A^{\ast}, B)$ symmetrically, each optimized with a reconstruction loss $\mathcal{L}_{\mathrm{rec}}$ for the graded output only, since no canonical ground truth exists in this phase. This step extends CanonCGT from discrete preset domains to continuous tonal and chromatic distributions, enabling adaptive color grading under diverse photographic conditions. Additional details, including loss terms, transformations $\mathcal{T}$, and hyperparameters, are provided in the supplement.

\section{Experiments}
\label{sec:exp}
We evaluate CanonCGT’s effectiveness and generalization quantitatively and qualitatively.

\subsection{Datasets}
CanonCGT is trained on both supervised and unsupervised datasets. The supervised dataset, derived from FiveK \cite{bychkovsky2011learning}, establishes a canonical prior and constructs the latent space for grading styles, while the unsupervised datasets support self-supervised generalization.

\medskip
\noindent\textbf{Supervised dataset:}
In the supervised phase, CanonCGT is trained on professionally retouched data providing consistent tone and color mappings. We use FiveK~\cite{bychkovsky2011learning}, which contains 5,000 photos retouched by five experts (A$\sim$E).
Among them, the expert~C edits are adopted as the canonical style set, as they exhibit neutral tonal balance and minimal stylistic bias. Prior studies~\cite{hu2018exposure,zeng2020learning,he2020conditional,chen2018deep,wang2019underexposed,moran2020deeplpf} have also identified expert~C as the most visually consistent and balanced reference among the experts, making this set well suited for defining the canonical domain.
From this set, 56 additional styles are synthesized using Lightroom presets --- 20 built-in and 36 user-generated --- each representing a distinct grading scheme. The 5,000 images are divided into 4,500 for training and 500 for validation across the 56 styles and resized to a shorter side of 480 pixels for efficiency.

\begin{table}[t]
\centering
\caption{Unsupervised datasets for self-supervised learning.}
\label{tab:unsupervised_datasets}
\vspace*{-0.2cm}
\setlength{\tabcolsep}{5pt}
\scriptsize
\begin{tabular}{lrrrl}
\toprule
Dataset & Total & Train & Test & Notes \\
\midrule
Flickr2K~\cite{lim2017enhanced} & 2,650 & 2,650 & 0 & High-quality natural photos \\
LSDIR~\cite{li2023lsdir} & 84,991 & 76,991 & 8,000 & Large-scale diverse scenes \\
PPR10K~\cite{liang2021ppr} & 11,161 & 8,875 & 2,286 & Expert~A version, portraits \\
DIV2K~\cite{agustsson2017ntire} & 900 & 0 & 900 & High-resolution benchmark \\
Food-101~\cite{bossard2014food} & 1,010 & 0 & 1,010 & Food photography \\
GLD-v2~\cite{weyand2020google} & 8,000 & 0 & 8,000 & Landmark scenes \\
\midrule
Total & 108,712 & 88,516 & 20,196 & \\
\bottomrule
\end{tabular}
\vspace*{-0.1cm}
\end{table}

\medskip
\noindent\textbf{Unsupervised datasets:}
After establishing the canonical prior, CanonCGT is further refined in the self-supervised phase to enhance robustness and generalization across diverse photographic domains. We use six public benchmarks covering a wide range of scenes, lighting conditions, and color distributions, as summarized in \Cref{tab:unsupervised_datasets}. In total, 88,516 images are used for training and 20,196 for testing. This large-scale setup enables CanonCGT to generalize beyond preset-defined styles, achieving faithful reference-based color grading across diverse image domains.

\begin{figure*}[!t]
\vspace*{-0.1cm}
\centering
\includegraphics[width=1\linewidth]{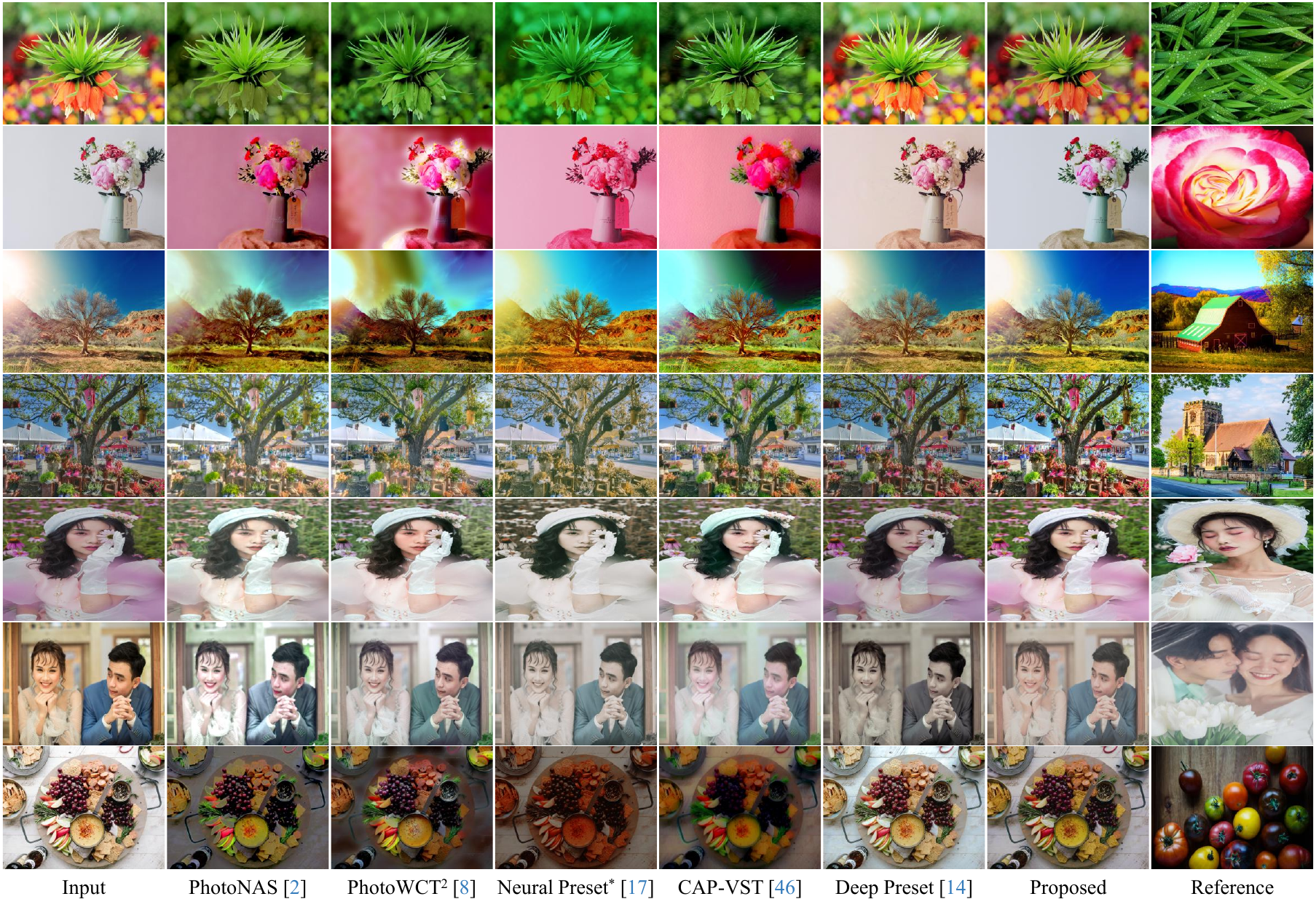}
\vspace*{-0.6cm}
\caption{Comparison of color grading results. CanonCGT produces natural and tonally balanced outputs, closely matching reference styles while preserving structural fidelity. Best viewed digitally with zoom for detailed comparison of tonal differences.}
\label{fig:Qualitative}
\vspace*{-0.3cm}
\end{figure*}

\subsection{Evaluation}\label{sec:eval}
Evaluation is conducted on the unsupervised test splits.

\medskip
\noindent\textbf{Protocols:}
Since no dataset provides paired supervision for reference-based color grading, we adopt a self-referential evaluation protocol consistent with the self-supervised training scheme. Each exemplar is divided into two non-overlapping regions, $A$ and $B$, sharing the same intrinsic tonal characteristics. A random color grading transformation $t \sim \mathcal{T}$ is applied to $A$ to obtain $A^{\ast}$, simulating real tonal variation. During evaluation, the model grades the perturbed input $A^{\ast}$ toward the original reference tone in region $B$, and the result is compared with the original region $A$. Unlike previous methods relying on unpaired statistics-based metrics, this protocol directly measures tonal fidelity and perceptual quality under realistic conditions.

\medskip
\noindent\textbf{Metrics:}
For quantitative evaluation, we assess three aspects: fidelity, content preservation, and grading style alignment. Fidelity reflects pixel-level accuracy, content preservation measures perceptual and structural consistency, and grading style alignment evaluates tone matching.

For fidelity, we use PSNR, SSIM, and $\Delta E_{ab}$, computed as the $L_2$ distance in CIE~LAB space. For content preservation, we adopt LPIPS~\cite{zhang2018unreasonable} and SSIM$_\textrm{ED}$~\cite{ke2023neural}, which measures SSIM between edge responses~\cite{soria2022ldc} to capture perceptual and edge fidelity. For grading style alignment, we compute color statistics, including histogram correlation (H-Corr) and chi-squared distance (H-Chi)~\cite{pass1997comparing}, quantifying tonal distribution similarity. All metrics are computed between each output and its target.

\subsection{Comparative assessment}
CanonCGT is compared with photorealistic and filter-based methods --- PhotoNAS \cite{an2020ultrafast}, PhotoWCT$^{2}$ \cite{chiu2022photowct2}, Neural Preset \cite{ke2023neural}, CAP-VST \cite{wen2023cap}, and Deep Preset \cite{ho2021deep} --- using official or reproduced implementations.

\medskip
\noindent\textbf{Quantitative comparison:}
\Cref{table:Quantitative} summarizes average performance on all unsupervised test datasets. CanonCGT consistently outperforms recent photorealistic and filter-based approaches \cite{an2020ultrafast,ho2021deep,chiu2022photowct2,ke2023neural,wen2023cap} across all metrics, achieving well-balanced color grading and preserving tonal fidelity and perceptual consistency. Per-dataset results in the supplement show strong performance across domains and also clear gains on DIV2K, Food-101, and GLD-v2 --- datasets used only for testing --- demonstrating reliable generalization beyond training data.

\medskip
\noindent\textbf{Qualitative comparison:}
\Cref{fig:Qualitative} compares color grading results. Existing methods often show excessive tonal shifts, texture distortions, or unstable color balance. In the 1st, 2nd, 5th, and 7th rows, where the references are relatively monotone, PhotoNAS, PhotoWCT$^{2}$, Neural Preset, and CAP-VST degrade color harmony, yielding unnatural tones. In the 3rd row, PhotoNAS, PhotoWCT$^{2}$, and CAP-VST also distort local textures and edges. Neural Preset alleviates such artifacts through deterministic color mapping, but its feature switching often introduces tonal inconsistency. Deep Preset produces stable tones yet frequently retains or accumulates the original style (1st, 3rd, 4th, 5th, and 7th rows). In contrast, CanonCGT generates perceptually balanced, coherent results by decoupling style removal and style adaptation in its two-stage canonicalization-grading framework. For example, CanonCGT faithfully transfers the vivid flower tone (2nd row), the soft mood of the blurred portrait (6th row), and the fruit coloration (last row), ensuring both tonal consistency and structural fidelity.

\begin{figure}[t]
    \centering
    \includegraphics[width=1\linewidth]{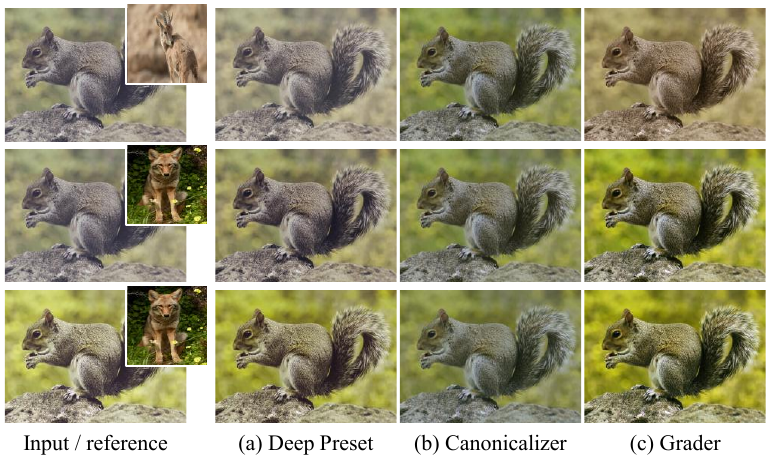}
    \vspace*{-0.6cm}
    \caption{Comparison with Deep Preset~\cite{ho2021deep}. (a) Deep Preset results, (b) CanonCGT canonicalized outputs, and (c) final graded results.} 
    \vspace*{-0.1cm}
\label{fig:vs_FST}
\end{figure}

\medskip
\noindent\textbf{Comparison with Deep Preset:}
\Cref{fig:vs_FST} highlights CanonCGT’s advantage over Deep Preset in style neutrality and consistency. In the first row, Deep Preset retains the hazy tone of the input, failing to match the reference. In the second and third rows, the inputs differ but share the same reference; Deep Preset carries over color bias from each input, leading to inconsistent results. In contrast, CanonCGT first canonicalizes the input into a tone-neutral form, as shown in column (b), and then applies the reference tone in a controlled manner. The final graded results in column (c) faithfully follow the reference and remain consistent across inputs, validating the stability of the canonicalization-grading framework.

\medskip
\noindent\textbf{User study:}
To complement quantitative evaluation, we conducted a user study under a cross-image setup similar to prior style-transfer work~\cite{yoo2019photorealistic,ho2021deep,ke2023neural}. Twenty-five participants with photography experience compared color-graded results from 50 randomly selected input-reference pairs from DIV2K, Food-101, and GLD-v2, yielding 1,250 trials. Each trial displayed the input, the reference, and results from CAP-VST, Deep Preset, and CanonCGT in random order. Participants evaluated two aspects: \textit{tonal consistency} with the reference, measuring how faithfully each result reproduces the target tone and color mood, and \textit{perceptual integrity}, assessing naturalness and structural fidelity. Lower ranks indicate higher quality. As summarized in \Cref{table:MOS}, CanonCGT achieves the lowest average ranks in both criteria, demonstrating consistent tonal alignment and superb perceptual fidelity.

\begin{table}[t]
    \centering
    \footnotesize
    \caption{Average ranking scores from 25 participants over 50 input-reference pairs (1,250 evaluations).}
    \label{table:MOS}
    \vspace{-0.2cm}
    \setlength{\tabcolsep}{6pt}
    \begin{tabular}{lcc}
    \toprule
    Method & Tonal consistency $\downarrow$ & Perceptual integrity $\downarrow$ \\
    \midrule
    CAP-VST      & 1.89 & 2.49 \\
    Deep Preset  & 2.33 & 1.97 \\
    CanonCGT     & \textbf{1.78} & \textbf{1.54} \\
    \bottomrule
    \end{tabular}
    \vspace*{-0.2cm}
\end{table}

\subsection{Analysis}\label{sec:analysis}

\noindent\textbf{Efficacy of the two-stage pipeline:}
We compare the two-stage CanonCGT with a one-stage variant that merges the canonicalizer and grader into a single LUT generator conditioned on concatenated input and reference grades. As shown in \Cref{table:pipeline}, the two-stage design performs better across all metrics, especially PSNR and $\Delta E_{ab}$, verifying that decoupling bias removal from grading ensures more stable tone mapping.

\medskip
\noindent\textbf{Efficacy of DP-CGT:}
We compare the model trained only in the supervised phase with the one further refined through the self-supervised phase. \Cref{table:analysis_SSL} shows that the second phase substantially improves performance on the unsupervised test datasets while maintaining comparable fidelity on the supervised validation set. This confirms that self-supervised phase enhances generalization and tonal stability under unconstrained style conditions.

\begin{figure}[t]
    \centering
    \includegraphics[width=1\linewidth]{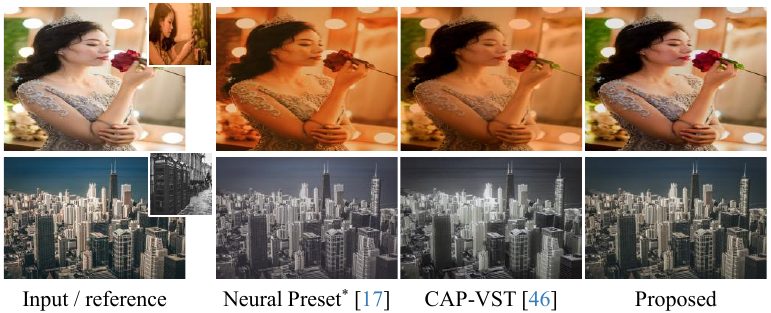}
    \vspace*{-0.6cm}
    \caption{Limitations of CanonCGT.} 
    \label{fig:limitations}
    \vspace*{-0.1cm}
\end{figure}

\begin{table}[!t]
    \centering
    \footnotesize
    \caption{Ablation of the two-stage CanonCGT design on the supervised validation set (FiveK val.).}
    \vspace*{-0.2cm}
    \begin{tabular}[t]{lcccc}
    \toprule
       & PSNR $\uparrow$ & SSIM $\uparrow$ & $\Delta E_{ab} \downarrow$ & LPIPS $\downarrow$ \\
    \midrule
        One-stage                       & 29.43           & 0.9366          &  4.53                      & 0.0678 \\
        Two-stage     & \textbf{30.29}  & \textbf{0.9405} &  \textbf{4.25}             & \textbf{0.0669} \\
    \bottomrule
    \end{tabular}
    \label{table:pipeline}
    \vspace*{-0.1cm}
\end{table}

\begin{table}[t]
    \centering
    \footnotesize
    \caption{Ablation of the dual-phase training framework, DP-CGT.}
    \vspace*{-0.2cm}
    \label{table:analysis_SSL}
    \setlength{\tabcolsep}{5pt}
    \begin{tabular}{lcccc}
    \toprule
    Dataset & \multicolumn{2}{c}{Supervised} & \multicolumn{2}{c}{Unsupervised} \\
    \cmidrule(l){2-3} \cmidrule(l){4-5}
    Metric & PSNR $\uparrow$ & $\Delta E_{ab} \downarrow$ & PSNR $\uparrow$ & $\Delta E_{ab} \downarrow$ \\
    \midrule
    Supervised only        & \textbf{30.29} & \textbf{4.25} & 19.17 & 15.07 \\
    + Self-supervised phase & 29.80 & 4.42 & \textbf{28.99} & \textbf{5.46} \\
    \bottomrule
    \end{tabular}
    \vspace*{-0.2cm}
\end{table}

\medskip
\noindent\textbf{Limitations:}
As shown in \Cref{fig:limitations}, when the reference exhibits a strong global tint, CanonCGT transfers the dominant hue selectively to preserve overall color balance. In the top row, with a reddish reference, it enhances the flower and mirror frames while keeping other regions natural, whereas competing methods apply the tint uniformly across the image. In the bottom row, with a black-and-white reference, CanonCGT preserves the dark sky tone rather than enforcing a full monochrome look, maintaining photorealism but limiting its ability to follow extreme reference styles.

\section{Conclusions}


We presented CanonCGT, a reference-based color grading framework built on a canonical pivot. By decomposing color grading into canonicalization and grading, CanonCGT removes intrinsic tonal bias and enables stable reference-driven tone mapping. The dual-phase training scheme (DP-CGT) combines supervised preset learning with self-supervised refinement on unconstrained photographs, improving robustness and generalization. Experiments demonstrate superior fidelity and perceptual consistency over existing photorealistic and filter-based methods. Beyond color grading, the canonicalization-grading formulation offers a new paradigm for learning style-aware image transformations.

\vspace{-0.15cm}
\section*{Acknowledgements}
\vspace{-0.15cm}
This work was supported by the National Research Foundation of Korea (NRF) funded by the Korea Government (MSIT) (No. RS-2024-00397293, RS-2022-NR068986, RS-2025-24523695), and by the AI Computing Infrastructure Enhancement (GPU Rental Support) User Support Program funded by MSIT (No. RQT-25-090187).

\clearpage

\clearpage
\appendix
\renewcommand{\thesection}{\Alph{section}}
\renewcommand{\thetable}{S-\arabic{table}}
\renewcommand{\thefigure}{S-\arabic{figure}}
\setcounter{section}{0}
\setcounter{table}{0}
\setcounter{figure}{0}

\twocolumn[{\centering\Large\bfseries Supplemental Materials \\[0.3em] CanonCGT: Reference-Based Color Grading via Canonical Pivot Representation\par}\vspace{1cm}]

\section{Implementation Details}\label{sec:Details}

We use a pretrained MobileNet-v2~\cite{sandler2018mobilenetv2} as the grade extractor and as the feature backbone for both the canonicalizer and the grader. The canonicalizer and the grader share the same network architecture (Fig.~\bu{4}) but do not share parameters. All models are optimized with AdamW~\cite{loshchilov2017decoupled} using $\beta_1=0.9$, $\beta_2=0.999$, and a weight decay of $5\times10^{-5}$. The learning rate starts at $2\times10^{-4}$ and decays to $5\times10^{-5}$ following a cosine annealing schedule~\cite{loshchilov2017sgdr}. We set $L=224$, $C=64$, and $N=17$, which provides a good balance between accuracy and computational cost.

\begin{figure}[t]
    \centering
    \includegraphics[width=1\linewidth]{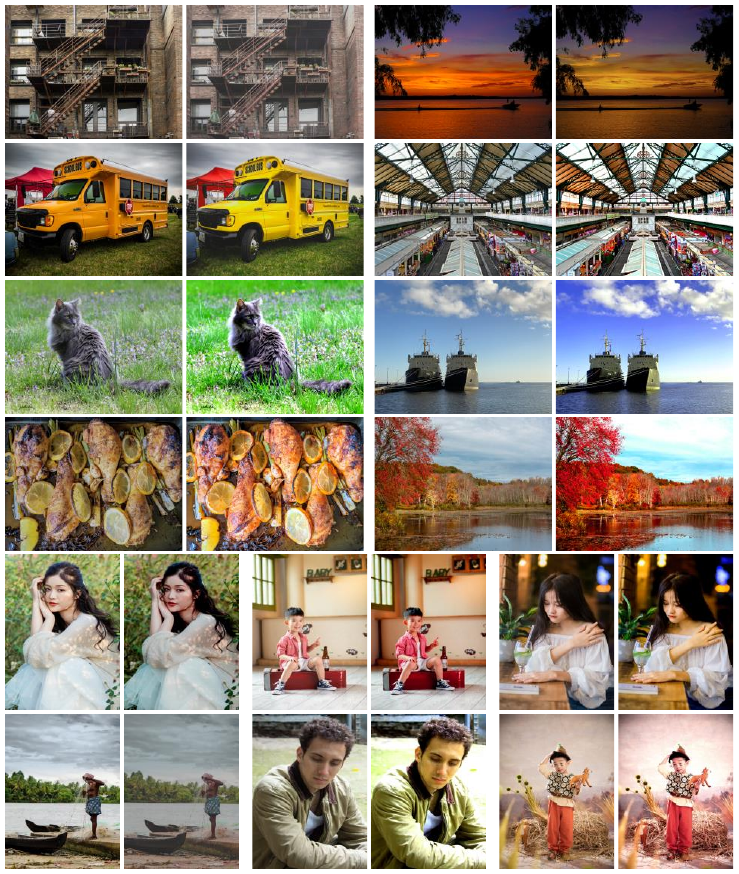}
\caption{Examples of color perturbations $t \sim \mathcal{T}$ used in the self-supervised training phase. For each pair, the left image is the original and the right is the perturbed version. These perturbations introduce perceptually diverse variations in mood, lighting, and color temperature.}
    \label{fig:augment}
    \vspace{-0.1cm}
\end{figure}

Training images are augmented by random cropping and horizontal flipping. In the self-supervised phase, we apply additional color perturbations --- brightness, contrast, saturation, and hue adjustments --- to simulate diverse tonal variations. Each crop is perturbed by a random transformation $t\sim\mathcal{T}$, where the adjustment factors are uniformly sampled within predefined ranges: $\pm 10\sim \pm 40\%$ for brightness, contrast, and saturation, and $\pm 2\sim \pm 5\%$ for hue. \Cref{fig:augment} shows examples of these perturbations.

All experiments run on a single NVIDIA RTX 3090 GPU. Full training of CanonCGT, including both supervised and self-supervised stages, takes roughly seven days.

\section{Training Details}

\subsection{Loss Functions}
To train CanonCGT, we employ two objectives: a supervised contrastive loss for the grade extractor and a reconstruction loss for the canonicalizer and the grader. The contrastive loss encourages compact and discriminative style embeddings, while the reconstruction loss enforces faithful tone reproduction.

\medskip
\noindent\textbf{Contrastive loss:} 
To construct a compact and discriminative latent space of grading styles, we adopt the supervised contrastive (SupCon) loss~\cite{khosla2020supervised}.

Consider a batch of $B$ images $\{X_i\}_{i=1}^{B}$ from the supervised dataset, where each image $X_i$ is synthesized with a preset $s_i$ and mapped to a grading vector $x_i$.
Positive and negative sets are defined according to the applied presets,  $\mathcal{P}(i)=\{j \mid s_j=s_i \}$ and $\mathcal{N}(i)=\{k \mid s_k\neq s_i \}$. Then, the SupCon loss is given by
\begin{align}
&\mathcal{L}_{\mathrm{supcon}}
= \notag \\
&-\frac{1}{B} 
   \sum_{i=1}^{B} 
   \frac{1}{|\mathcal{P}(i)|}
   \sum_{j\in\mathcal{P}(i)}
   \log 
   \frac{
        \exp(x_i^{\top} x_j / \tau)
   }{
        \sum_{k\in\mathcal{N}(i)}
        \exp(x_i^{\top} x_k / \tau)
   } ,
\label{eq:SupCon}
\end{align}
where $\tau$ is the temperature.

\medskip
\noindent\textbf{Reconstruction loss:} 
The canonicalizer and the grader are trained with a multi-term reconstruction loss 
\begin{equation}
    \mathcal{L}_{\rm{rec}} = w_{m} \mathcal{L}_{m} + w_{g} \mathcal{L}_{g} + w_{p} \mathcal{L}_{p},
    \label{eq:rec_loss}
\end{equation}
where $\mathcal{L}_{m}$, $\mathcal{L}_{g}$, and $\mathcal{L}_{p}$ denote the MSE loss, the gradient loss~\cite{ma2020structure}, and the perceptual loss~\cite{zhang2018unreasonable}, respectively. These terms jointly enforce pixel-wise accuracy, gradient consistency, and perceptual similarity. We set $w_{m}=1$, $w_{g}=1$, and $w_{p}=0.05$.

The MSE loss $\mathcal{L}_{m}$ between an output image $\hat{I}$ and its target $\tilde{I}$ is defined as the mean squared error. The gradient loss $\mathcal{L}_{g}$ penalizes discrepancies in horizontal and vertical gradients, 
\begin{equation}
    \mathcal{L}_{g} = \lVert \nabla_{x} \hat{I} - \nabla_{x} \tilde{I} \rVert_{2} + \lVert \nabla_{y} \hat{I} - \nabla_{y} \tilde{I} \rVert_{2}.
\end{equation}
The perceptual loss $\mathcal{L}_{p}$ measures the difference in a deep feature space, 
\begin{equation}
    \mathcal{L}_{p} = \sum_{k\in\{4,9,16,23\}} \lVert f_{k}(\hat{I}) - f_{k}(\tilde{I}) \rVert_{1},
\end{equation}
where $f_{k}(\cdot)$ denotes the activation of the $k$-th layer of VGG-16~\cite{simonyan2014very} pretrained on ImageNet~\cite{russakovsky2015imagenet}.

\subsection{Optimization}
CanonCGT is trained using the dual-phase color grading training (DP-CGT) scheme described in the main paper, consisting of a supervised warm-up phase followed by a self-supervised refinement phase.

\medskip
\noindent\textbf{Supervised warm-up phase:}
In this phase, the grade extractor, canonicalizer, and grader are trained sequentially and then jointly to establish a stable canonical prior.
The grade extractor is first trained with the supervised contrastive loss $\mathcal{L}_{\mathrm{supcon}}$ in \eqref{eq:SupCon}.
After its convergence, the extractor parameters are frozen, and the canonicalizer and grader are optimized using the reconstruction loss $\mathcal{L}_{\mathrm{rec}}$ in \eqref{eq:rec_loss}.
Finally, all modules are fine-tuned together in an end-to-end manner under the combined objectives, with weights of 0.05, 1, and 1 assigned to the grade extractor, canonicalizer, and grader, respectively. Batch sizes are set to 64 for the grade extractor and 16 for the canonicalizer and grader during their individual training, and 9 during the end-to-end fine-tuning stage.

\medskip
\noindent\textbf{Self-supervised refinement phase:}
Building on the canonical prior, CanonCGT is further refined through the self-referential reconstruction described in the main paper.
Because the overall tone of unlabeled photographs is not well-defined, the grade extractor is frozen in this phase, and only the canonicalizer and grader are updated using the reconstruction loss $\mathcal{L}_{\mathrm{rec}}$, applied to the graded output only, as no canonical supervision is available.
Training is performed on both the supervised and unsupervised datasets, with each batch composed of 9 samples: 6 drawn from the unsupervised set and 3 from the supervised set.

\section{Canonical Style}\label{sec:supp_Canonical}
We analyze how the choice of the canonical style set influences CanonCGT. For a controlled comparison, only the canonical style set is varied, while all other training configurations remain identical to those in the main paper. Four canonical style sets are evaluated:

\begin{itemize}
\item \textit{Expert C (Proposed)}: Images retouched by expert~C in FiveK, exhibiting neutral tonal balance and minimal stylistic bias.
\item \textit{Preset-biased (Warm-Contrast)}: Expert~C images regraded with a warm, high-contrast Lightroom preset that increases color temperature and mid-tone contrast.
\item \textit{Preset-biased (Cool-Matte)}: Expert~C images regraded with a cool, desaturated preset that reduces chroma and shifts the white balance toward bluish tones.
\item \textit{Preset-mean (56-style)}: The per-image mean appearance computed over all 56 preset-defined styles, serving as a bias-reduced reference domain.
\end{itemize}
The two preset-biased variants impose opposite tonal biases (warm versus cool), while the preset-mean variant provides a style-neutral reference by averaging appearance across all preset-defined styles.

\begin{table}[t]
    \centering
    \footnotesize
    \caption{Comparison of canonical style sets evaluated on the supervised validation split of FiveK.}
    \begin{tabular}{lccc}
    \toprule
    Canonical style set             & PSNR & SSIM & $\Delta E_{ab}$ \\
    \midrule
    Expert C (Proposed)             & 30.29 & 0.9405 & 4.25     \\
    Preset-biased (Warm-Contrast)   & 25.82 & 0.8923 & 5.96     \\
    Preset-biased (Cool-Matte)      & 26.71 & 0.9041 & 5.58     \\
    Preset-mean (56-style)          & 29.80 & 0.9375 & 4.32     \\
    \bottomrule
    \end{tabular}
    \label{table:canonical}
\end{table}

As shown in \Cref{table:canonical}, the two preset-biased variants exhibit noticeably lower fidelity, indicating that strong stylistic shifts compromise the neutrality of the canonical domain and lead to less consistent grading behavior. In contrast, the preset-mean variant performs on par with expert~C, suggesting that averaging multiple style variants effectively suppresses preset-specific biases and yields a stable, neutral canonical domain suitable for CanonCGT.

\section{Component Analysis}\label{sec:Component}
We analyze the three key components of CanonCGT --- the grade extractor, canonicalizer, and grader --- to understand how the learned grading representations and the LUT-based modules jointly contribute to reliable color grading.

\subsection{Grade Extractor}
\begin{figure}[t]
    \centering
    \includegraphics[width=1\linewidth]{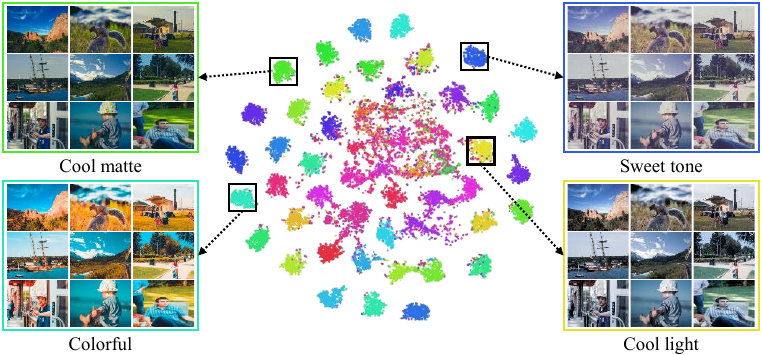}
    \caption{ t-SNE visualization~\cite{maaten2008visualizing} of grading vectors on the FiveK validation set. Each cluster corresponds to a distinct preset-defined style, with example images illustrating representative tonal characteristics.} 
    \label{fig:t-SNE}
\end{figure}

\noindent\textbf{Tone-aware latent space:}
The grade extractor encodes color grading styles into a compact and discriminative latent space. As shown in \Cref{fig:t-SNE}, the t-SNE visualization \cite{maaten2008visualizing} on the FiveK validation set reveals clear clusters aligned with the 56 preset-defined styles, indicating that the supervised contrastive objective organizes the embedding space according to tonal semantics.

\medskip
\noindent\textbf{Style decoding accuracy:}
To quantify the separability of the learned embeddings, we evaluate $k$-NN decoding accuracy of styles. For each test embedding, we retrieve its $k$ nearest neighbors and assign a predicted style by majority voting over their labels. As shown in \Cref{table:classification}, the decoding accuracy remains consistently high across different values of $k$, demonstrating that the grading vectors retain strong style-discriminative information.

Because the grade extractor is frozen during the self-supervised refinement phase, its decoding performance remains unchanged from the supervised stage, as expected. The embedding space therefore remains fixed and is used consistently by the canonicalization and grading modules.

\medskip
\noindent\textbf{Intra-style coherence and inter-style separation:}
We analyze cosine similarities between grading vectors to examine how the embedding space captures intra-style coherence and inter-style separation. Three evaluation modes are employed:
(i)~\textit{Self-referential}, between cropped views of the same image;
(ii)~\textit{Intra-style}, between different images edited with the same preset; and
(iii)~\textit{Inter-style}, between images with different presets.
For the self-referential case, each image is randomly cropped multiple times, and similarities are averaged for stable measurement.

As shown in \Cref{table:Content_invariance}, the similarities exhibit a clear hierarchy --- highest for self-referential pairs, slightly lower for intra-style pairs, and much lower for inter-style pairs. This trend indicates that the embedding space preserves strong self-consistency while providing meaningful discrimination between styles. The high self-referential similarity also aligns with the assumption in the self-supervised phase that two cropped views of the same exemplar share the same underlying grading style.

\begin{figure*}[!t]
    \centering
    \includegraphics[width=1\linewidth]{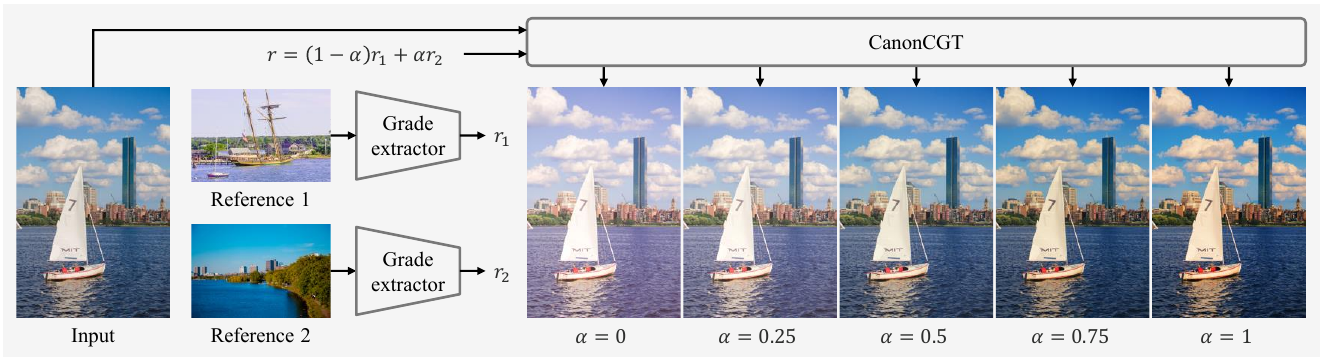}
\caption{Continuous transition across grading styles. Two reference images provide grading vectors $r_1$ and $r_2$, which are linearly interpolated to obtain $r$. Conditioned on this interpolated vector, CanonCGT generates smooth tonal transitions between the two styles, demonstrating the continuity of the learned grading space.}
    \label{fig:interpolation}
\end{figure*}

\medskip
\noindent\textbf{Continuity of the learned grading space:}
\Cref{fig:interpolation} illustrates that CanonCGT produces smooth tonal transitions when conditioned on interpolated grading vectors between two reference styles. This behavior suggests that the embedding space forms a semantically continuous manifold rather than exhibiting discrete jumps across styles. Such continuity ensures stable responses for reference images whose grading vectors lie between well-defined style categories and supports reliable behavior during the self-supervised refinement phase, where the model encounters unpaired and diverse images.

\begin{table}[t]
    \centering
    \footnotesize
    \caption{$k$-NN decoding accuracy of the grade embeddings on the FiveK validation set.}
    \begin{tabular}{lccc}
        \toprule
        & $k=5$ & $k=10$ & $k=20$ \\
        \midrule
        Decoding accuracy (\%) & 84.47 & 83.01 & 82.33 \\
        \bottomrule
    \end{tabular}
    \label{table:classification}
\end{table}

\begin{table}[t]
    \centering
    \footnotesize
    \caption{Intra-style coherence and inter-style separation of grade embeddings.}
    \begin{tabular}{lcc}
        \toprule
             & Mean cosine similarity & Relative scale \\
        \midrule
        Self-referential    & 0.8881 & $\times$1.00 \\
        Intra-style         & 0.8533 & $\times$0.96 \\
        Inter-style         & 0.3421 & $\times$0.39 \\
        \bottomrule
    \end{tabular}
    \label{table:Content_invariance}
\end{table}

\subsection{Canonicalizer and Grader}
CanonCGT consists of two key components: the canonicalizer, which removes the intrinsic tonal bias of the input, and the grader, which applies a target grading vector to synthesize the final styled output.

\medskip
\noindent\textbf{Canonicalizer.}
We assess canonical reconstruction fidelity on the FiveK validation set, where each preset-styled image is paired with its expert~C retouching as the canonical ground truth. As shown in \Cref{table:Canonicalizer}, the supervised warm-up phase delivers stable reconstruction quality, while the subsequent self-supervised refinement introduces a slight decrease in PSNR, SSIM, and $\Delta E_{ab}$. This is expected, as the refinement phase prioritizes broader generalization rather than strict canonical reconstruction.

\medskip
\noindent\textbf{Grader.}
For grading evaluation, each canonical input $A$ is matched to its ground-truth preset retouching $A_S$. We use CLIP-based retrieval to locate the most similar expert~C exemplar $B$, and extract the reference grading vector from its preset-styled version $B_S$. Conditioned on this vector, the grader synthesizes the styled output for $A$, which is compared against $A_S$. As shown in \Cref{table:Grader}, the supervised warm-up phase achieves accurate reconstruction under controlled style conditions, while the refinement stage yields slightly lower PSNR, SSIM, and $\Delta E_{ab}$.

\medskip
\noindent\textbf{Discussion.}
Although the refinement phase reduces supervised reconstruction accuracy for both modules, it substantially improves robustness under diverse photographic conditions, as demonstrated in Table~\bu{5} of the main paper. These results indicate that DP-CGT enhances generalization and stabilizes the overall behavior of CanonCGT.

\section{Ablations}\label{sec:supp_Analysis}
\Cref{table:Ablations} analyzes the effect of key architectural choices on the supervised validation set (FiveK val.). Each experiment varies a single factor while keeping all other components fixed, allowing us to examine how model capacity influences color fidelity and tonal accuracy.

\begin{table}[t]
    \centering
    \footnotesize
\caption{Reconstruction fidelity of the canonicalizer on the FiveK validation set.}
    \begin{tabular}{lccc}
        \toprule
        Phase & PSNR $\uparrow$ & SSIM $\uparrow$ & $\Delta E_{ab} \downarrow$ \\
        \midrule
        Supervised (warm-up)        & \textbf{28.47} & \textbf{0.8957} & \textbf{5.658} \\
        Self-supervised (refined)   & 27.35 & 0.8931 & 6.063 \\
        \bottomrule
    \end{tabular}
    \label{table:Canonicalizer}
\end{table}

\begin{table}[t]
    \centering
    \footnotesize
\caption{Reconstruction fidelity of the grader on the FiveK validation set.}
    \begin{tabular}{lccc}
        \toprule
        Phase & PSNR $\uparrow$ & SSIM $\uparrow$ & $\Delta E_{ab} \downarrow$ \\
        \midrule
        Supervised (warm-up)        & \textbf{30.79} & \textbf{0.9271} & \textbf{4.379} \\
        Self-supervised (refined)   & 29.65 & 0.9224 & 4.503 \\
        \bottomrule
    \end{tabular}
    \label{table:Grader}
\end{table}

\medskip
\noindent\textbf{Number of query tokens ($N^3$):}  
We first evaluate the impact of the number of query tokens. The LUT sampling grid forms an $N \times N \times N$ lattice, yielding $N^3$ tokens that determine the sampling density of both the canonical and grading LUT spaces. As shown in \Cref{table:Ablations}, performance improves with larger $N$, reflecting the benefit of finer sampling for color transformation. However, because the number of tokens grows cubically with $N$ and the quadratic cost of cross-attention increases rapidly with token count, we adopt $N=17$ as a balanced configuration that provides sufficient precision without excessive overhead.

\begin{table}[t]
    \centering
    \footnotesize
\caption{Ablation of key architectural choices of CanonCGT on the FiveK validation set. The proposed configuration is highlighted in light gray.}

    \begin{tabular}{llccc}
    \toprule
    Setting &                       & PSNR & SSIM & $\Delta E_{ab}$ \\
    \midrule
    \multirow{3}[0]{*}{Number of query tokens ($N^3$)} 
        & $9^3$                         & 29.87 & 0.9389 & 4.38 \\
        & \cellcolor{gray!15}$17^3$     & \cellcolor{gray!15}30.29 & \cellcolor{gray!15}0.9405 & \cellcolor{gray!15}4.25 \\
        & $33^3$                        & \textbf{30.31} & \textbf{0.9407} & \textbf{4.23} \\
    \midrule
    \multirow{3}[0]{*}{Number of channels ($C$)} 
        & 32                        & 29.74 & 0.9372 & 4.46 \\
        & \cellcolor{gray!15}64     & \cellcolor{gray!15}30.29 & \cellcolor{gray!15}0.9405 & \cellcolor{gray!15}4.25 \\
        & 96                        & \textbf{30.32} & \textbf{0.9408} & \textbf{4.24} \\
    \midrule
    \multirow{3}[0]{*}{Number of attention heads} 
        & 2                         & 29.68 & 0.9376 & 4.43 \\
        & \cellcolor{gray!15}4      & \cellcolor{gray!15}\textbf{30.29} & \cellcolor{gray!15}\textbf{0.9405} & \cellcolor{gray!15}\textbf{4.25} \\
        & 8                         & 30.11 & 0.9394 & 4.28 \\
    \midrule
    \multirow{3}[0]{*}{Number of enc./dec. blocks} 
        & 4/6                       & 29.72 & 0.9379 & 4.42 \\
        & \cellcolor{gray!15}6/8    & \cellcolor{gray!15}\textbf{30.29} & \cellcolor{gray!15}0.9405 & \cellcolor{gray!15}\textbf{4.25} \\
        & 8/10                      & 30.17 & \textbf{0.9410} & 4.27 \\
    \bottomrule
    \end{tabular}
    \label{table:Ablations}
\end{table}

\begin{table*}
    \centering
    \footnotesize
\caption{Comparison of GPU runtime, memory usage, and model size. All results are measured with Float32 precision on an NVIDIA RTX~3090. The numbers in parentheses indicate the exact image resolution. Units: ``s'' for seconds, ``GB'' for gigabytes, and ``M'' for millions of parameters. ``OOM'' indicates an out-of-memory failure.}
    \begin{tabular}[t]{+L{2.6cm}^C{2.6cm}^C{2.6cm}^C{2.6cm}^C{2.6cm}^C{1.8cm}}
    \toprule
    \multirow{2}[5]{*}{Method}      
        & \multicolumn{4}{c}{GPU runtime $\downarrow$ / Memory $\downarrow$}                                        & Model size $\downarrow$   \\ 
        \cmidrule(l){2-5}       \cmidrule(l){6-6}
        & \makecell[c]{FHD\\$(1920 \times 1080)$} & \makecell[c]{2K\\$(2560 \times 1440)$} & \makecell[c]{4K\\$(3840 \times 2160)$} & \makecell[c]{8K\\$(7680 \times 4320)$}    & \makecell[r]{Number of \\parameters} \\
    \midrule
        PhotoNAS~\cite{an2020ultrafast}             & 0.537\,s\,/\,15.60GB     & 0.908\,s\,/\,23.87GB     & OOM                      & OOM  & \makecell[r]{40.24\,M}   \\
        PhotoWCT$^{2}$~\cite{chiu2022photowct2}     & 0.277\,s\,/\,14.09GB     & 0.424\,s\,/\,19.75GB     & 1.016\,s\,/\,23.79GB     & OOM  & \makecell[r]{7.05\,M}   \\
        Neural Preset$^{*}$ \cite{ke2023neural}  & 0.013\,s\,/\;\;\,0.27GB  & 0.016\,s\,/\;\;\,0.42GB  & 0.019\,s\,/\;\;\,0.88GB  & 0.037\,s\,/\,3.38GB  & \makecell[r]{4.66\,M}   \\
        CAP-VST~\cite{wen2023cap}                   & 0.462\,s\,/\;\;\,4.68GB  & 0.871\,s\,/\;\;\,9.96GB  & 2.116\,s\,/\,21.83GB & OOM & \makecell[r]{\textbf{4.09\,M}}   \\
        Deep Preset~\cite{ho2021deep}               & 0.323\,s\,/\;\;\,8.81GB  & 0.437\,s\,/\,13.21GB     & 1.114\,s\,/\,22.68GB     & OOM  & \makecell[r]{267.77\,M}  \\
    \midrule
        Proposed                    & \textbf{0.008\,s\,/\;\;\,0.26GB}   & \textbf{0.009\,s\,/\;\;\,0.42GB}    & \textbf{0.010\,s\,/\;\;\,0.82GB}    & \textbf{0.017\,s\,/\,3.04GB}  & \makecell[r]{5.01\,M}   \\
    \bottomrule
    \end{tabular}
    \label{table:Complexities}
\end{table*}

\medskip
\noindent\textbf{Channel dimension ($C$):}  
We next examine the effect of the channel dimension, which governs the representational capacity of CanonCGT. Increasing $C$ from 32 to 64 consistently improves all metrics, indicating that a wider feature space captures color and tone relationships more effectively. Further increasing $C$ to 96 yields only marginal gains in PSNR and $\Delta E_{ab}$ while incurring additional parameter and memory cost. We therefore adopt $C=64$ as a well-balanced setting.

\medskip
\noindent\textbf{Number of attention heads:}  
We also analyze how the number of attention heads influences performance. Using only two heads limits cross-channel interaction and slightly reduces fidelity, whereas increasing to eight heads provides only marginal improvement while increasing computational complexity. We thus choose four heads as an effective and efficient configuration.

\medskip
\noindent\textbf{Encoder–decoder depth:}  
Lastly, we vary the numbers of encoder and decoder blocks to assess the effect of network depth. A shallower structure (4/6) lacks sufficient modeling capacity to capture fine tonal variations, resulting in lower accuracy. A deeper configuration (8/10) produces only minor gains in SSIM but introduces additional latency and memory usage. The proposed depth of six encoder and eight decoder blocks achieves the best trade-off.

\section{More Results}

\subsection{Complexities}
\Cref{table:Complexities} summarizes the computational efficiency of CanonCGT across resolutions from FHD to 8K. GPU runtime, memory usage, and model size are measured on a machine equipped with an AMD Ryzen 9 3900X CPU and a single NVIDIA RTX 3090 GPU. All values are averaged over 100 inference iterations using identical I/O pipelines and batch configurations.

CanonCGT performs color grading with remarkable efficiency while maintaining high fidelity and tonal stability. The model requires only 0.26\,GB of memory and 8\,ms for FHD inference, and scales consistently to 8K resolution with a modest increase in runtime and memory usage. This efficiency primarily stems from its LUT-based formulation, which applies global color mapping rather than spatially dense convolutional operations. Overall, CanonCGT provides a practical solution suitable for real-time, high-resolution applications.

\begin{table*}
    \centering
    \footnotesize
    \caption{Quantitative results on the unsupervised datasets.}
    \begin{tabular}[t]{+L{1.0cm}^R{1.2cm}^C{1.7cm}^C{1.8cm}^C{2.3cm}^C{1.8cm}^C{2.0cm}^C{1.5cm}}
    \toprule
        Dataset & \makecell[c]{Metric} & PhotoNAS~\cite{an2020ultrafast} & PhotoWCT$^{2}$~\cite{chiu2022photowct2} &  Neural Preset$^{*}$ \cite{ke2023neural} & CAP-VST~\cite{wen2023cap} & Deep Preset~\cite{ho2021deep} & Proposed\\
    \midrule
    \multirow{7}{*}{DIV2K}
        & PSNR $\uparrow$                   & 17.35  & 17.23  & 19.10  & 18.95  & 18.86  & \textbf{29.25}     \\
        & SSIM $\uparrow$                   & 0.7452 & 0.8188 & 0.8461 & 0.8107 & 0.8571 & \textbf{0.9572}    \\
        & $\Delta E_{ab}$ $\downarrow$      & 18.30  & 18.39  & 16.25  & 16.96 & 14.76  & \textbf{5.43}      \\
        & LPIPS $\downarrow$                & 0.2921 & 0.1924 & 0.1972 & 0.2013 & 0.1492 & \textbf{0.0580}    \\
        & SSIM$_\textrm{ED}$ $\uparrow$     & 0.6538 & 0.7400 & 0.7176 & 0.7207 & 0.7646 & \textbf{0.8954}    \\
        & H-Corr $\uparrow$                 & 0.2695 & 0.3514 & 0.2913 & 0.3601 & 0.2727 & \textbf{0.5154}    \\
        & H-Chi $\downarrow$                & 0.2964 & 0.2500 & 0.3164 & 0.2495 & 0.2896 & \textbf{0.1603}    \\
    \midrule
    \multirow{7}{*}{PPR10K}
        & PSNR $\uparrow$                   & 15.81  & 14.80  & 17.06  & 16.48  & 17.99 & \textbf{29.31}     \\
        & SSIM $\uparrow$                   & 0.7423 & 0.7863 & 0.8379 & 0.7771 & 0.8696 & \textbf{0.9589}    \\
        & $\Delta E_{ab}$ $\downarrow$      & 20.41  & 22.04  & 19.16  & 20.41  & 15.44  & \textbf{4.98}      \\
        & LPIPS $\downarrow$                & 0.3613 & 0.2951 & 0.2871 & 0.3148 & 0.2220 & \textbf{0.1062}    \\
        & SSIM$_\textrm{ED}$ $\uparrow$     & 0.6584 & 0.7257 & 0.7243 & 0.6935 & 0.7669 & \textbf{0.8834}    \\
        & H-Corr $\uparrow$                 & 0.1565 & 0.2111 & 0.1656 & 0.2156 & 0.1688 & \textbf{0.4314}    \\
        & H-Chi $\downarrow$                & 0.3932 & 0.3444 & 0.4403 & 0.3655 & 0.3880 & \textbf{0.2310}    \\
    \midrule
    \multirow{7}{*}{LSDIR}
        & PSNR $\uparrow$                   & 17.23  & 17.01  & 19.11  & 18.72  & 18.85  & \textbf{29.37}     \\
        & SSIM $\uparrow$                   & 0.7476 & 0.8199 & 0.8457 & 0.8102 & 0.8546 & \textbf{0.9586}    \\
        & $\Delta E_{ab}$ $\downarrow$      & 18.83  & 19.12  & 16.56  & 17.74  & 14.96  & \textbf{5.33}      \\
        & LPIPS $\downarrow$                & 0.2935 & 0.1986 & 0.1993 & 0.2068 & 0.1512 & \textbf{0.0537}    \\
        & SSIM$_\textrm{ED}$ $\uparrow$     & 0.6551 & 0.7383 & 0.7175 & 0.7187 & 0.7618 & \textbf{0.8990}    \\
        & H-Corr $\uparrow$                 & 0.2898 & 0.3600 & 0.3052 & 0.3669 & 0.2967 & \textbf{0.5364}    \\
        & H-Chi $\downarrow$                & 0.2980 & 0.2552 & 0.3223 & 0.2553 & 0.2916 & \textbf{0.1624}    \\
    \midrule
    \multirow{7}{*}{Food-101}
        & PSNR $\uparrow$                   & 16.68  & 16.44  & 18.77  & 18.09  & 18.53  & \textbf{27.50}     \\
        & SSIM $\uparrow$                   & 0.7201 & 0.7965 & 0.8428 & 0.7982 & 0.8417 & \textbf{0.9444}    \\
        & $\Delta E_{ab}$ $\downarrow$      & 20.92  & 21.20  & 18.95  & 19.94  & 17.78  & \textbf{7.33}      \\
        & LPIPS $\downarrow$                & 0.3763 & 0.2743 & 0.2687 & 0.2600 & 0.2310 & \textbf{0.0972}    \\
        & SSIM$_\textrm{ED}$ $\uparrow$     & 0.5726 & 0.6717 & 0.6577 & 0.6632 & 0.6726 & \textbf{0.8262}    \\
        & H-Corr $\uparrow$                 & 0.2589 & 0.3347 & 0.3192 & 0.3351 & 0.3132 & \textbf{0.5517}    \\
        & H-Chi $\downarrow$                & 0.3113 & 0.2616 & 0.3143 & 0.2726 & 0.3032 & \textbf{0.1630}    \\
    \midrule
    \multirow{7}{*}{GLD-v2}
        & PSNR $\uparrow$                   & 16.38  & 16.06  & 18.18  & 17.59  & 18.54  & \textbf{28.69}     \\
        & SSIM $\uparrow$                   & 0.7427 & 0.8153 & 0.8467 & 0.8100 & 0.8608 & \textbf{0.9659}    \\
        & $\Delta E_{ab}$ $\downarrow$      & 19.62  & 20.22  & 17.38  & 18.96  & 15.12  & \textbf{5.48}      \\
        & LPIPS $\downarrow$                & 0.3240 & 0.2305 & 0.2246 & 0.2374 & 0.1812 & \textbf{0.0650}    \\
        & SSIM$_\textrm{ED}$ $\uparrow$     & 0.6316 & 0.7397 & 0.7229 & 0.7138 & 0.7605 & \textbf{0.8986}    \\
        & H-Corr $\uparrow$                 & 0.2456 & 0.3075 & 0.2769 & 0.3089 & 0.2796 & \textbf{0.5264}    \\
        & H-Chi $\downarrow$                & 0.3503 & 0.3094 & 0.3667 & 0.3160 & 0.3307 & \textbf{0.1837}    \\
    \bottomrule
    \end{tabular}
    \label{table:Quantitative_per}
\end{table*}

\subsection{Quantitative Comparisons}
\Cref{table:Quantitative_per} presents quantitative results on the unsupervised test datasets, including DIV2K, PPR10K, LSDIR, Food-101, and GLD-v2. For each dataset, we evaluate fidelity (PSNR, SSIM, $\Delta E_{ab}$), content preservation (LPIPS, SSIM$_\mathrm{ED}$), and style alignment (H-Corr, H-Chi) to comprehensively assess the quality of the graded outputs.

CanonCGT consistently outperforms existing photorealistic and filter-based methods, achieving substantial gains in fidelity and perceptual consistency, with higher PSNR and SSIM and lower $\Delta E_{ab}$ and LPIPS. Improvements in H-Corr and H-Chi further indicate that CanonCGT produces tonally aligned results and demonstrates strong generalization and robustness under diverse photographic conditions.

\subsection{Qualitative Comparisons}

\begin{figure*}
    \centering
    \includegraphics[width=1\linewidth]{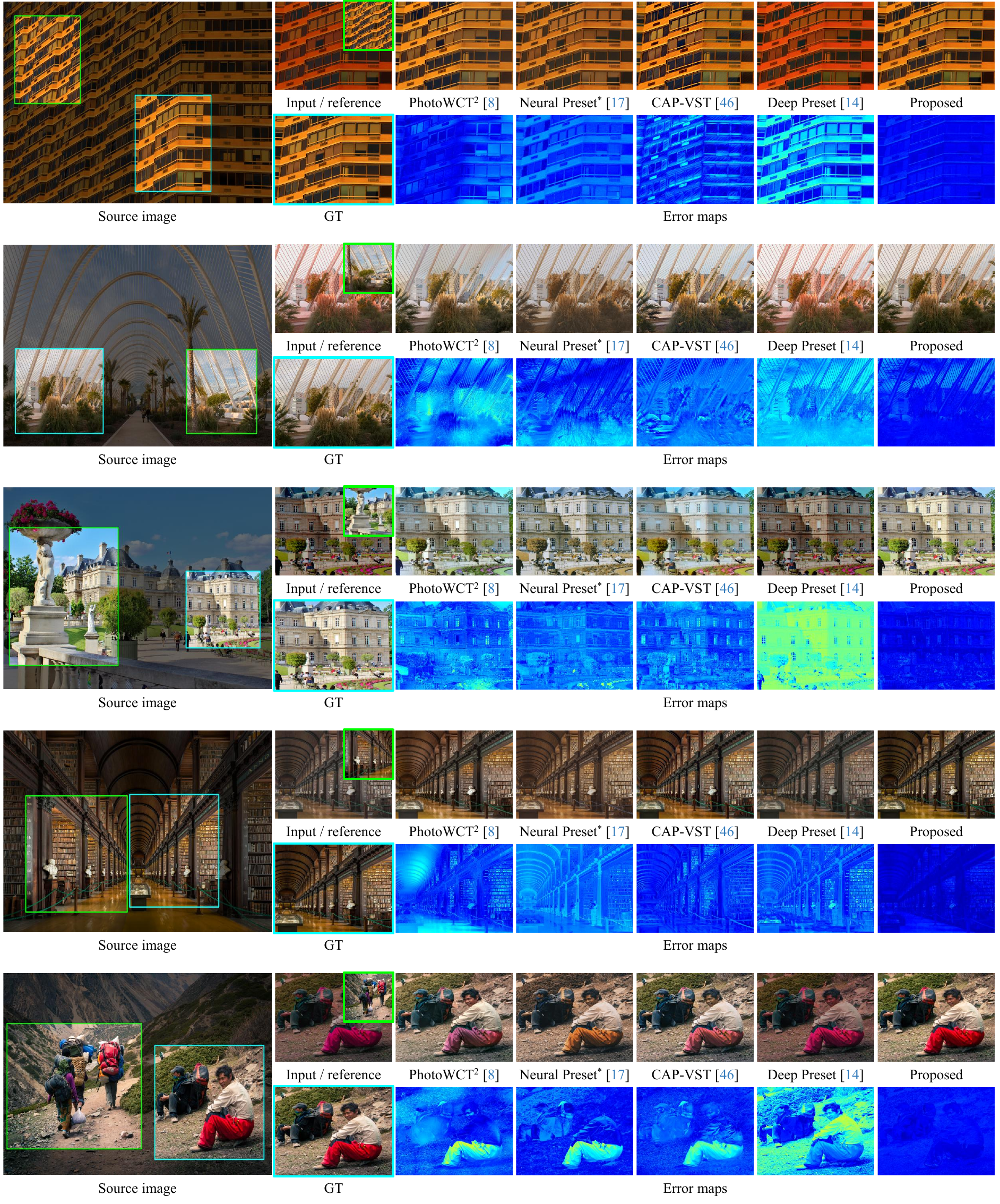}
    \vspace*{-0.7cm}
\caption{Qualitative comparison under the self-referential protocol on the unsupervised test set. For each sample, the green and cyan boxes denote the reference and ground-truth (GT) images, respectively; the input is generated by applying a color perturbation $t \sim \mathcal{T}$ to the GT. Results from competing methods~\cite{ho2021deep,chiu2022photowct2,ke2023neural,wen2023cap} and CanonCGT are shown with their corresponding error maps below.}
    \label{fig:self_referential_1}
\end{figure*}

\begin{figure*}
    \centering
    \includegraphics[width=1\linewidth]{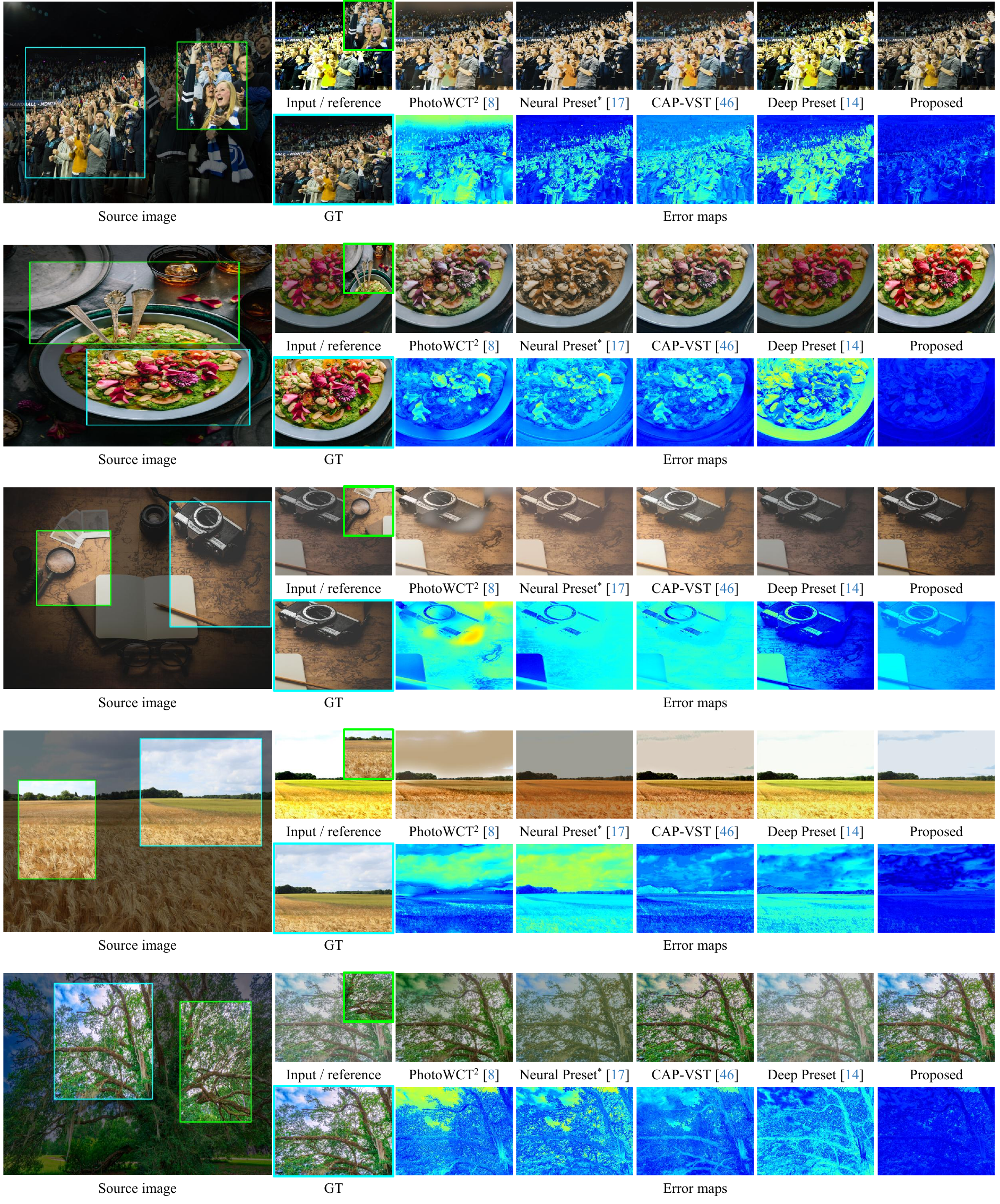}
    \caption{Qualitative comparison under the self-referential protocol on the unsupervised test set.}
    \label{fig:self_referential_2}
\end{figure*}

\begin{figure*}
    \centering
    \includegraphics[width=1\linewidth]{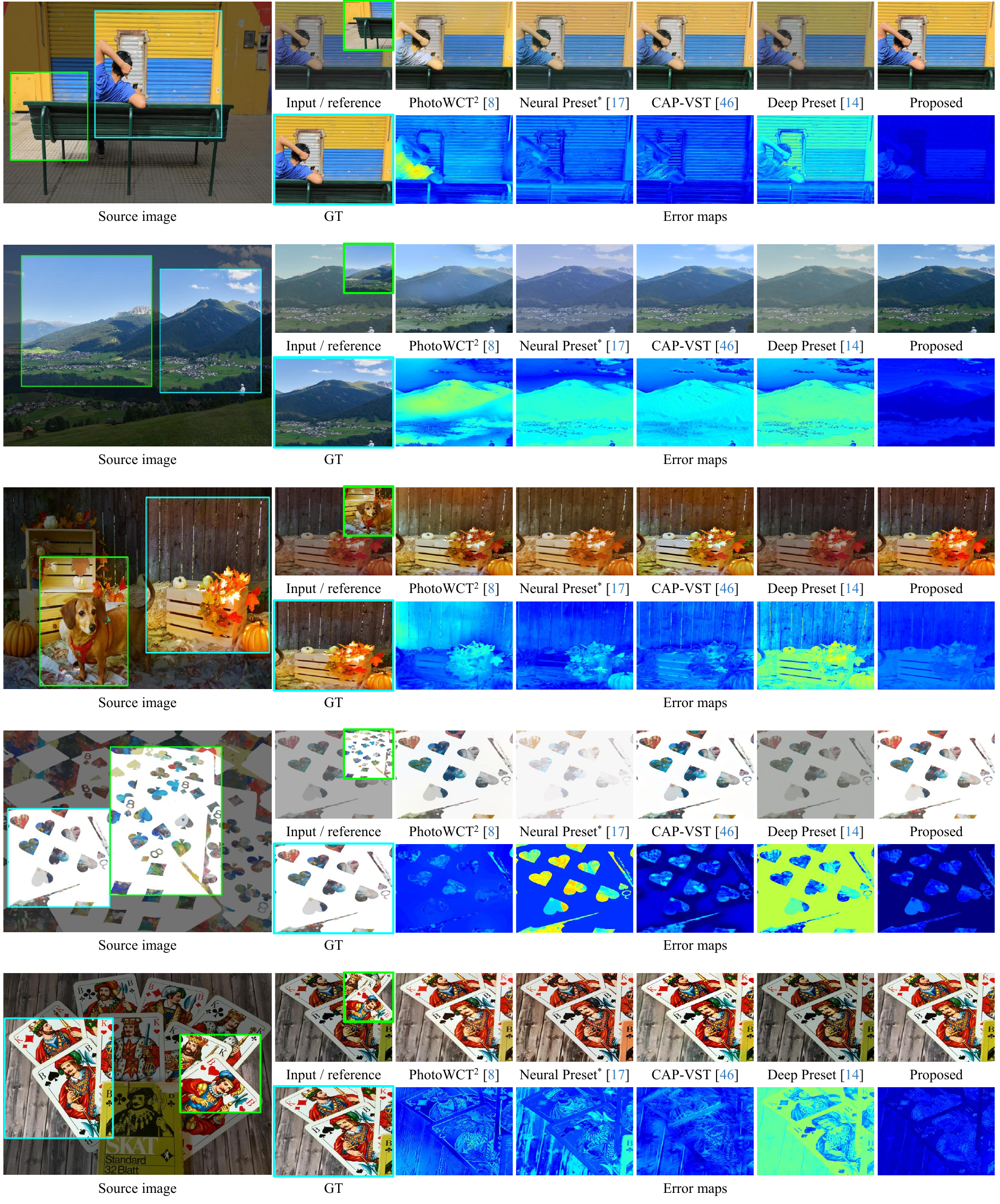}
    \caption{Qualitative comparison under the self-referential protocol on the unsupervised test set.}
    \label{fig:self_referential_3}
\end{figure*}

\begin{figure*}
    \centering
    \includegraphics[width=1\linewidth]{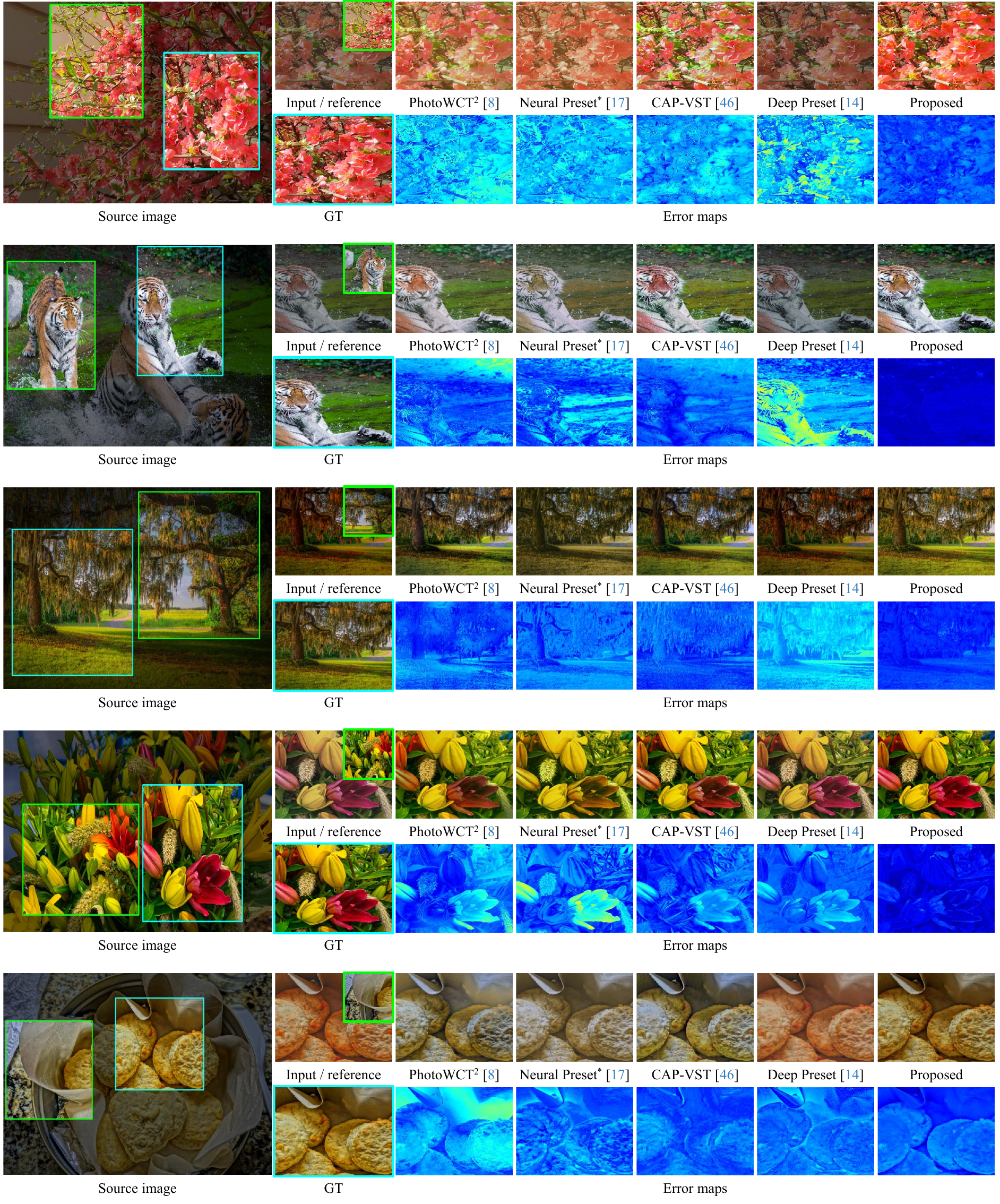}
    \caption{Qualitative comparison under the self-referential protocol on the unsupervised test set.}
    \label{fig:self_referential_4}
\end{figure*}

\noindent\textbf{Self-referential protocol:}
\Cref{fig:self_referential_1,fig:self_referential_2,fig:self_referential_3,fig:self_referential_4} show qualitative results under the self-referential protocol, where the input and reference come from the same exemplar. Competing methods~\cite{ho2021deep,chiu2022photowct2,ke2023neural,wen2023cap} often introduce local tonal inconsistencies or regionally unbalanced colors, whereas CanonCGT closely matches the reference tone, lighting, and color temperature. These results align with the quantitative gains on the unsupervised datasets (see \Cref{table:Quantitative_per} and Table~\bu{1} in the main paper), demonstrating CanonCGT’s ability to stabilize tonal bias under self-referential conditions.

\begin{figure*}
    \centering
    \includegraphics[width=1\linewidth]{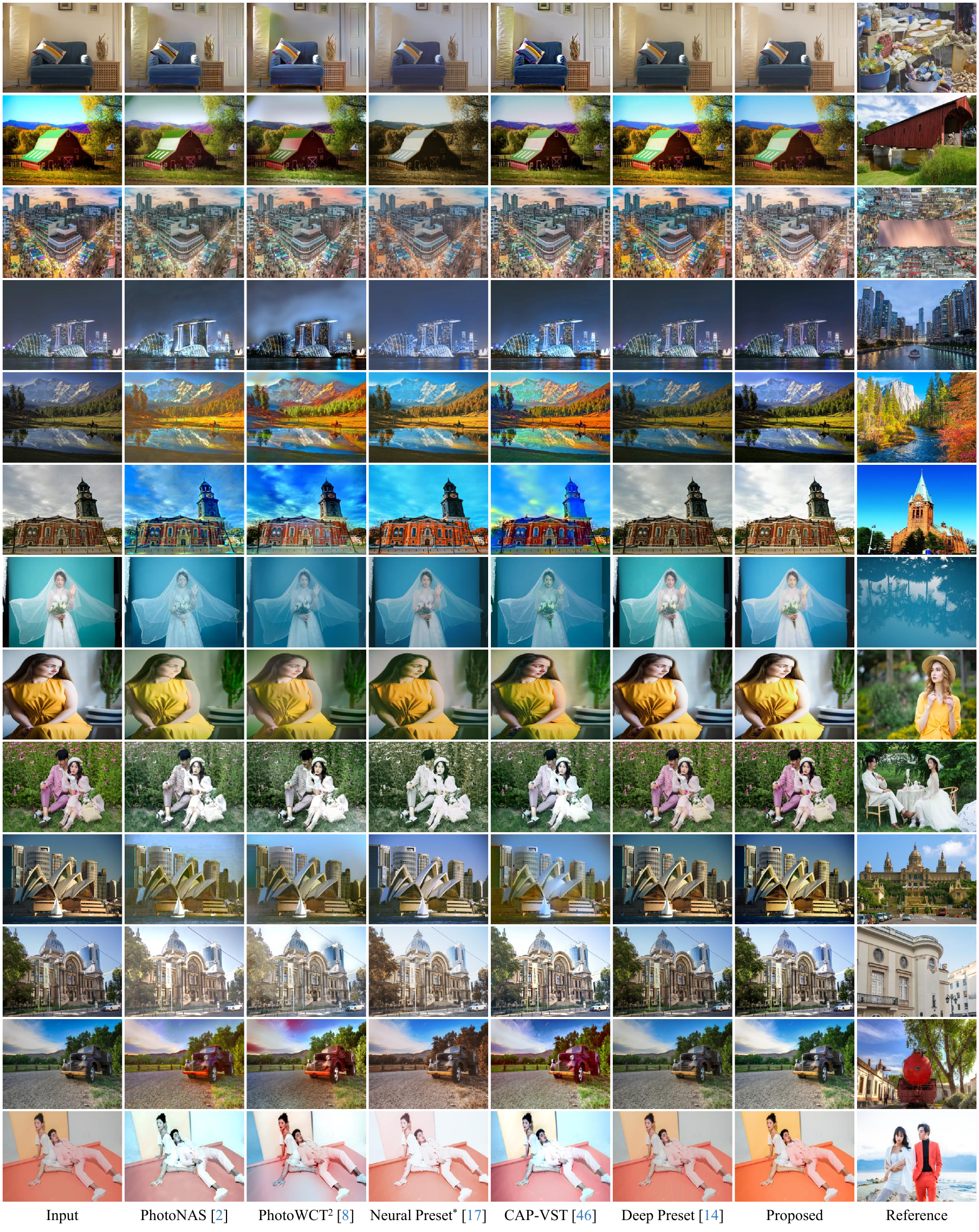}
    \vspace*{-0.7cm}
    \caption{Qualitative comparison under the cross-image protocol on the unsupervised test set.}
    \label{fig:cross_image_1}
\end{figure*}

\begin{figure*}
    \centering
    \includegraphics[width=1\linewidth]{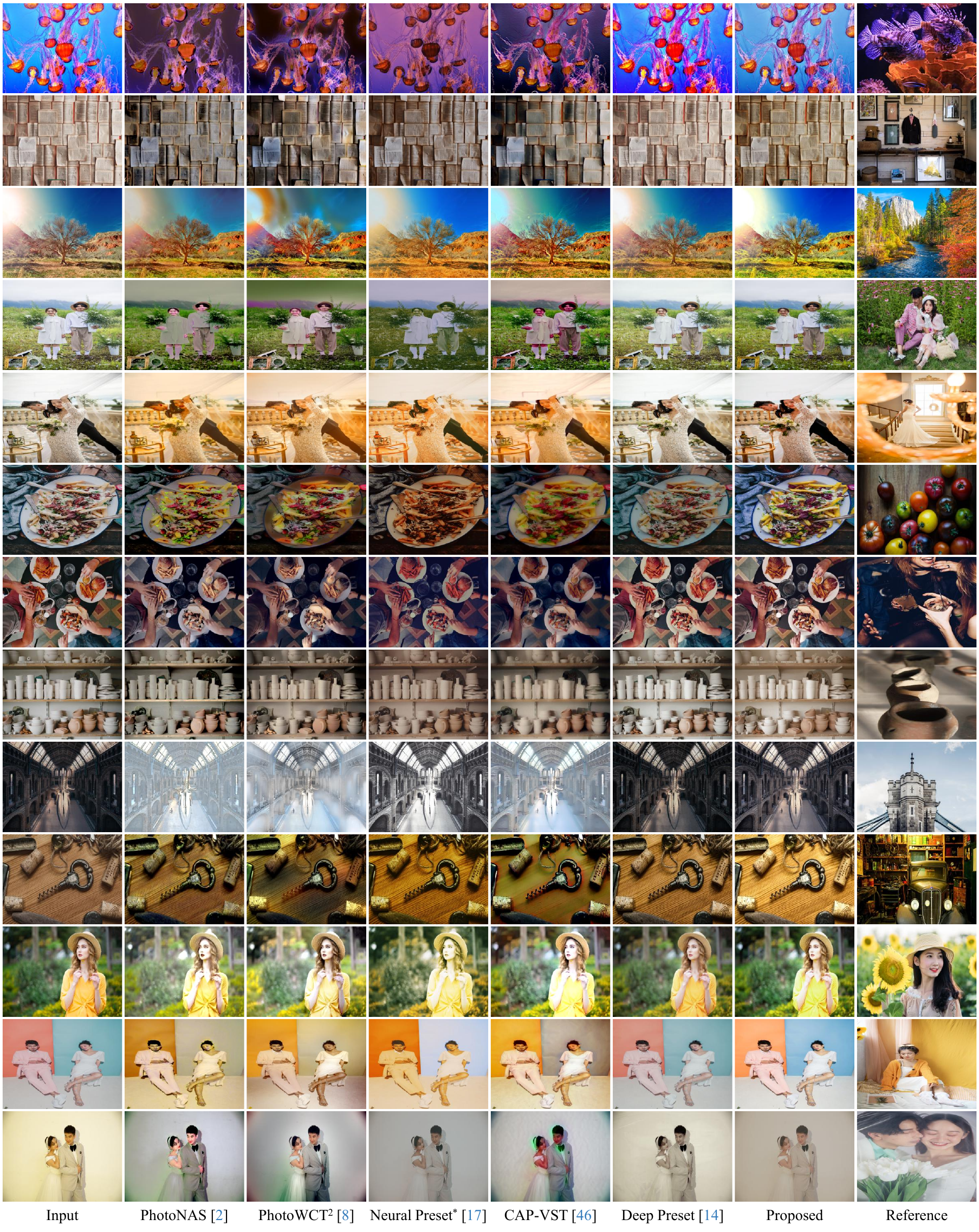}
    \vspace*{-0.7cm}
    \caption{Qualitative comparison under the cross-image protocol on the unsupervised test set.}
    \label{fig:cross_image_2}
\end{figure*}

\medskip
\noindent\textbf{Cross-image protocol:}
\Cref{fig:cross_image_1,fig:cross_image_2} present additional examples under the cross-image protocol. Deep Preset~\cite{ho2021deep} often retains the input’s original tonal bias, while other photorealistic methods~\cite{chiu2022photowct2,ke2023neural,wen2023cap} may produce local artifacts or inconsistent color transitions. CanonCGT reliably transfers the target tone while preserving structural realism and global color harmony, yielding coherent results across diverse scenes.

\subsection{Additional Results of CanonCGT}
\begin{figure*}[t]
    \centering
    \includegraphics[width=1\linewidth]{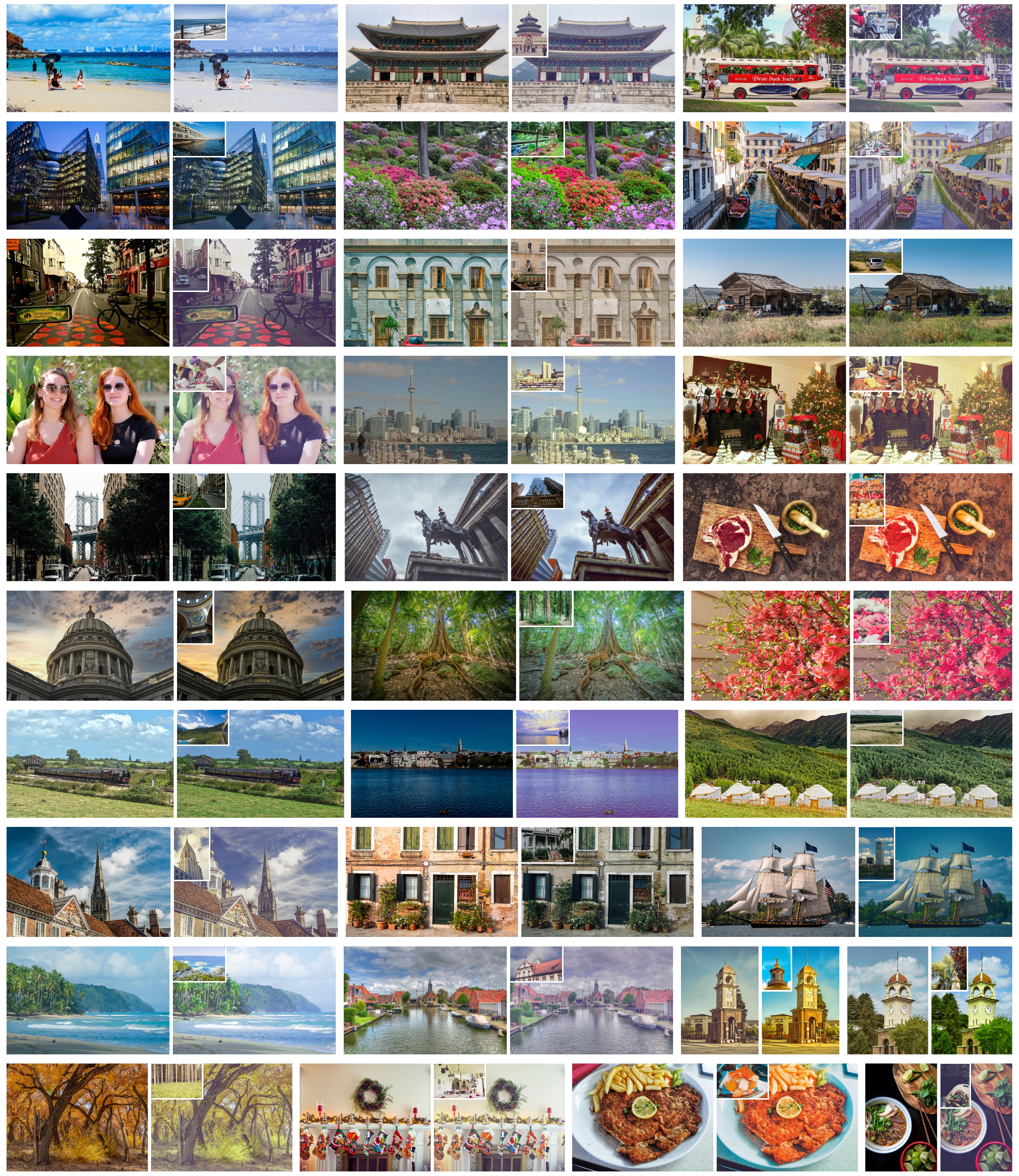}
\caption{Additional qualitative results of CanonCGT. For each pair, the left image shows the input and the right image shows the color-graded output using the inset reference image. CanonCGT produces photorealistic color grading that matches the tonal mood, lighting, and color temperature of the reference while preserving color harmony and scene structure.}
    \label{fig:more_result}
\end{figure*}

Finally, \Cref{fig:more_result} presents additional examples across diverse input–reference pairs, showing that CanonCGT produces consistent and photorealistic color grading with stable tonal behavior. The results demonstrate that the model preserves color harmony and scene structure while adapting effectively to a wide range of styles and scene conditions.

\clearpage
\clearpage
{\small
\bibliographystyle{ieeenat_fullname}
\bibliography{31756_arXiv}
}

\end{document}